\documentclass{article}

\usepackage[nonatbib,preprint]{neurips_2021}

\usepackage[utf8]{inputenc}
\usepackage[T1]{fontenc}
\usepackage{hyperref}
\usepackage{url}
\usepackage{booktabs}
\usepackage{amsmath}
\usepackage{amssymb}
\usepackage{amsfonts}
\usepackage{xfrac}
\usepackage{bm}
\usepackage{microtype}
\usepackage{graphicx}
\usepackage{wrapfig}
\usepackage{caption}
\usepackage{subcaption}
\usepackage{xcolor}

\graphicspath{{img/}}

\newcommand{\TITLE}{Multivariate Deep Evidential Regression}

\title{\TITLE}

\author{%
    Nis Meinert \\
    German Aerospace Center (DLR) \\
    \texttt{nis.meinert@dlr.de}
    \And
    Alexander Lavin \\
    Institute for Simulation Intelligence \\
    \texttt{lavin@simulation.science}
}

\begin{document}
\maketitle

\begin{abstract}
    There is significant need for principled uncertainty reasoning in machine learning systems as they are increasingly deployed in safety-critical domains.
    A new approach with uncertainty-aware neural networks (NNs), based on learning evidential distributions for aleatoric and epistemic uncertainties, shows promise over traditional deterministic methods and typical Bayesian NNs, yet several important gaps in the theory and implementation of these networks remain.
    We discuss three issues with a proposed solution to extract aleatoric and epistemic uncertainties from regression-based neural networks.
    The approach derives a technique by placing evidential priors over the original Gaussian likelihood function and training the NN to infer the hyperparameters of the evidential distribution.
    Doing so allows for the simultaneous extraction of both uncertainties without sampling or utilization of out-of-distribution data for univariate regression tasks.
    We describe the outstanding issues in detail, provide a possible solution, and generalize the deep evidential regression technique for multivariate cases.
\end{abstract}


\section{Introduction}
Using neural networks (NNs) for regression tasks is one of the main applications of modern machine learning.
Given a dataset of $(\vec{x}_i, \vec{y}_i)$ pairs, the typical objective is to train a NN $f(\vec{x}_i | \bm{w})$ w.r.t.\ $\bm{w}$ such that a given loss $\mathcal{L}(\vec{x}_i, \vec{y}_i)$ becomes minimal for each $(\vec{x}_i, \vec{y}_i)$ pair.
Traditional regression-based NNs are designed to output the regression target, a.k.a., the prediction for $\vec{y}_i$, directly which allows a subsequent minimization, for example of the sum of squares:
\begin{equation}
    \min\limits_{\bm{w}} \sum\limits_i \mathcal{L}(\vec{x}_i, \vec{y}_i) = \min\limits_{\bm{w}} \sum\limits_i  \underbrace{\left( \vec{y}_i - f(\vec{x}_i | \bm{w}) \right)^2}_{\mathcal{L}_i(\bm{w})} \,.
\end{equation}
Technically, this is nothing but a fit of a model $f$, parameterized with $\bm{w}$, w.r.t.\ $\sum_i \mathcal{L}_i$ to data.
As with any fit, the model has to find a balance between being too specific (over-fitting) and being too general (under-fitting).
In machine learning this balance is typically evaluated by analyzing the trained model on a separated part of the given data which was not seen during training.
In practice, no model will be able to describe this evaluation sample perfectly and deviations can be categorized into two groups: \textit{aleatoric} and \textit{epistemic} uncertainties~\cite{kendall17}. The former quantifies system stochasticity such as observation and process noise, and the latter is model-based or subjective uncertainty due to limited data.

In the following we will describe and analyze an approach to reliably estimate these kinds of uncertainties for NNs by modifying the architecture and introducing an appropriate loss function.
The structure of this paper is as follows:
First, we  will briefly discuss aleatoric and epistemic uncertainties using a pseudo example.
We then give an overview of the proposed solution of Amini et\ al.~\cite{amini20}.
In Sec.~\ref{sec:issues} we describe several issues with the prior work, and follow with a possible solution in Sec.~\ref{sec:solution}.
Finally, in Sec.~\ref{sec:multi} we summarize our multivariate generalization approach extending the prior work, which we use throughout the text.

\subsection{Aleatoric and epistemic uncertainty}
\begin{wrapfigure}{r}{.45\textwidth}
    \centering
    \includegraphics[scale=.8]{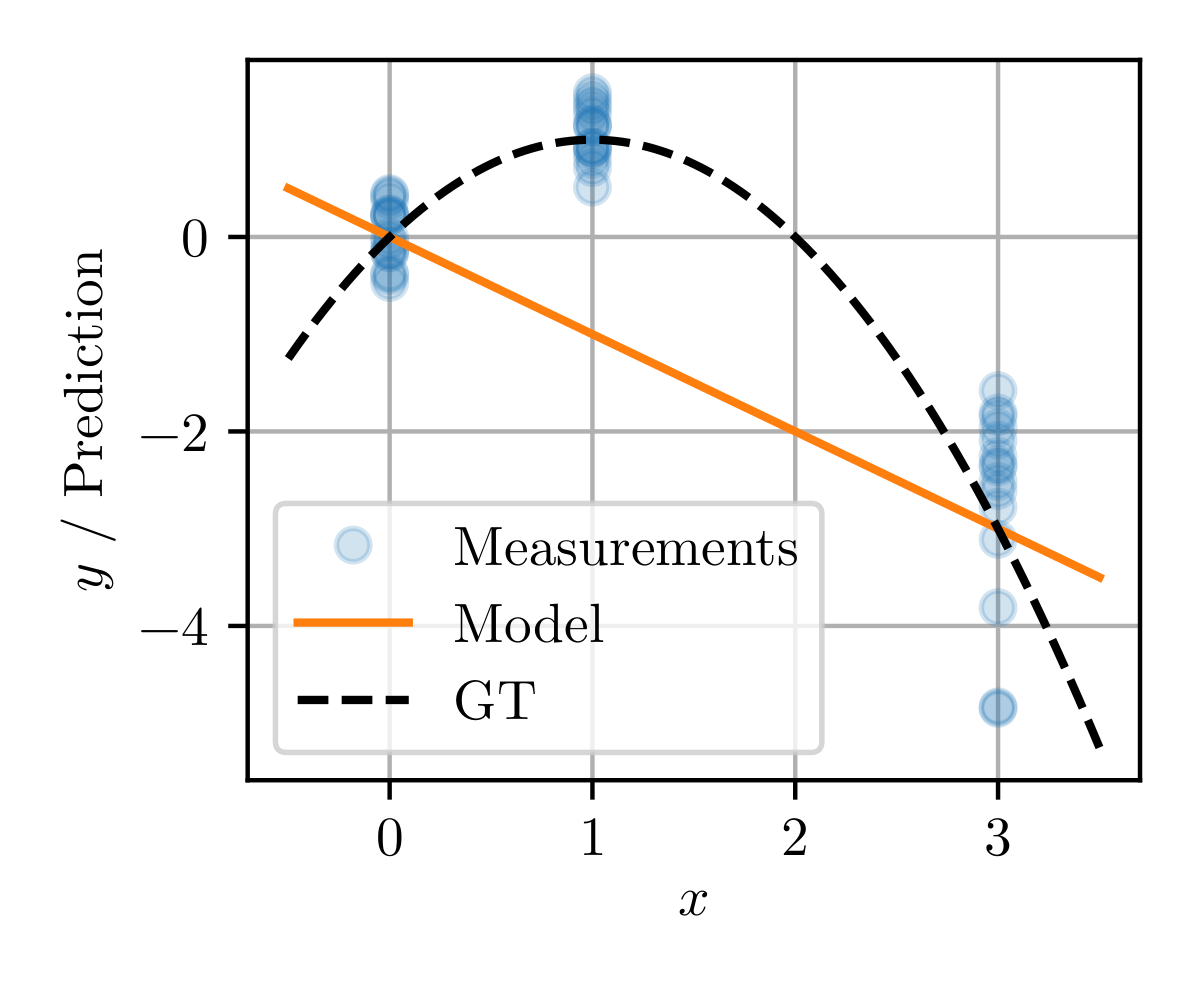}
    \caption{A fit of a model to toy data. The model has low aleatoric uncertainty at $x=\{0,1,2\}$ and large aleatoric uncertainty at $x=3$, whereas the epistemic uncertainty is low at $x=\{0,3\}$ and large elsewhere.}
    \label{fig:unctypes}
\end{wrapfigure}
In Fig.~\ref{fig:unctypes} we show data located at $x=\{0, 1, 3\}$ where for each value of $x_i$ multiple measurements, $\vec{y}_i$, were taken.
We generated these data by sampling from a normal distribution centered at the dashed line referred to as the ground truth (GT).
The model (solid line) represents the prediction for different values of $x_i$.
The uncertainty in data is low for $x=\{0,1\}$ and large for $x=3$, leading to a low aleatoric uncertainty at the former points and a high aleatoric uncertainty at the latter where there is high variance in the observed data.
Similarly, the epistemic uncertainty is low at $x=\{0,3\}$ where predictions are close to the observed data, and large at $x=1$ where the model poorly fits the observed data.
In general, aleatoric uncertainty is related to the noise level in the data and, typically, does not depend on the sample size -- only the shape of this uncertainty becomes sharper with increasing sample size.
In contrast, epistemic uncertainty does scale with the sample size, and either allows the model to be pulled towards the observed distribution at $x=1$ if only the sample size in this region is increased, or allows the fit of a more complex model and thus decreasing under-fitting in general.

Pivotal for this work is the point $x=2$ (and, technically, all other points $\mathbb{Z} \setminus \{0, 1, 3\}$) since no data were observed here.
Having no data also corresponds to an epistemic uncertainty since it decreases, in theory, if more data are drawn, assuming a conclusive data sample.
In contrast to the large epistemic uncertainty at $x=1$ this uncertainty is hard to detect by evaluating a trained model, but at the same time it can be crucial for models to communicate this type of uncertainty in real-world applications such as autonomous driving~\cite{geiger12,bojarski16,godard17}, where models can easily be confronted with out-of-distribution data that was underrepresented during training, leading to dangerous and expensive failures.

\subsection{Deep Evidential Regression}
Different approaches have been developed to enable models to estimate either aleatoric or epistemic uncertainty, where the latter often requires out-of-distribution data or compute-intense sampling, limiting the application of such approaches~\cite{malinin18,gal16,lakshminarayanan17,jain21}.
Recently, Amini et~al.\ adopted a technique from the classification realm and attacked this problem by changing the interpretation of the parameters of the NN~\cite{amini20,sensoy18}:
The number of output neurons of a NN for a univariate regression task has to be increased from one to four.
The output of these neurons are interpreted as $\alpha$, $\beta$, $\mu_0$, and $\kappa \in \mathbb{R}$.%
\footnote{In~\cite{amini20} the authors refer to them as $\alpha$, $\beta$, $\gamma$ and $\nu$, respectively.}
These are the parameters of a normal-inverse-gamma function $\operatorname{\texttt{NIG}}(\mu_0, \kappa;\, \alpha, \beta)$, and used to estimate the prediction and both uncertainties as:
\begin{equation}
    \label{eq:uncs}
    \underbrace{\mathbb{E}[\mu] = \mu_0}_{\text{prediction}}
    \qquad\qquad
    \underbrace{\mathbb{E}[\sigma^2] = \beta / (\alpha - 1)}_{\text{aleatoric}}
    \qquad\qquad
    \underbrace{\operatorname{var}[\mu] = \mathbb{E}[\sigma^2] / \kappa}_{\text{epistemic}}
\end{equation}
The authors derive these relations by taking the normal-inverse-gamma distribution (\texttt{NIG}) as the conjugated prior of a normal distribution with unknown mean $\mu$ and standard deviation $\sigma$.
Further, the authors show that by using Bayesian inference the loss function for these four parameters becomes a scaled Student's $t$-distribution with $2\alpha$ degrees of freedom (DoF), parametrized as:
\begin{equation}
    \mathcal{L}_i^\text{NLL} = -\log \operatorname{St}_{2\alpha}\!\left( y_i \middle| \mu_0, \frac{\beta(1 + \kappa)}{\kappa \alpha} \right).
\end{equation}
For reasons we will discuss later, they combine it with a second loss function, referred to as the \textit{evidence regularizer}, using the \textit{total evidence} $\Phi$, yielding the total loss:
\begin{equation}
    \mathcal{L}_i(\bm{w}) = \mathcal{L}_i^\text{NLL}(\bm{w}) + \lambda \times |y_i - \mu_0| \Phi \,,
\end{equation}
where the coupling, $\lambda$, is a hyperparameter of the model.
Note that, following the notation of~\cite{amini20}, we have dropped indices for all parameters for the sake of brevity -- see Appendix~\ref{sec:gfit} for a more elaborated discussion.

\section{Addressing issues in the prior art}
\label{sec:issues}
In this section we discuss three issues with the prior work on Deep Evidential Regresstion~\cite{amini20}.
We also describe new multivariate formulations,  which will be detailed later in Sec.~\ref{sec:multi}.

\subsection{Definition of total evidence}
In Bayesian inference a normal-inverse-Wishart distribution (\texttt{NIW}) is a conjugate prior for i.i.d.\ drawn events from a multivariate normal distribution with unknown mean $\vec{\mu} \in \mathbb{R}^n$ and unknown variance $\bm{\Sigma} \in \mathbb{R}^{n \times n}$~\cite{gelman04}.
In the univariate case, $n=1$, a \texttt{NIW} distribution becomes a \texttt{NIG} distribution and we slightly change our notation and use $\mu \in \mathbb{R}$ and $\sigma^2 \in \mathbb{R}$ for the (unknown) mean and the (unknown) variance, respectively.
These prior densities,
\begin{align}
    p(\mu, \sigma^2) &= \texttt{NIG}(\mu_0, \kappa;\, \alpha, \beta) &
    p(\vec{\mu}, \bm{\Sigma}) &= \texttt{NIW}(\vec{\mu}_0, \kappa;\, \bm{\Psi}, \nu)
\end{align}
with\footnote{As eluded previously we suppress indices for the sake of brevity.} $\mu_0, \kappa, \alpha, \beta, \nu \in \mathbb{R}$, $\vec{\mu}_0 \in \mathbb{R}^n$ and $\bm{\Psi} \in \mathbb{R}^{n \times n}$, correspond to the assumption that each pair $(\mu, \sigma^2)$ or $(\vec{\mu}, \bm{\Sigma})$ is sampled from a normal distribution, $\mathcal{N}$, and an inverse gamma distribution, $\Gamma^{-1}$, or inverse Wishart distribution, $\mathcal{W}^{-1}$,
\begin{subequations}
    \begin{align}
        \sigma^2 &\sim \Gamma^{-1}(\alpha, \beta) &
        \bm{\Sigma} &\sim \mathcal{W}^{-1}(\bm{\Psi}, \nu) \equiv \mathcal{W}^{-1}(\nu \bm{\Sigma}_0, \nu) \\
        \mu | \sigma^2 &\sim \mathcal{N}(\mu_0, \sigma^2 / \kappa) &
        \vec{\mu} | \bm{\Sigma} &\sim \mathcal{N}(\vec{\mu}_0, \bm{\Sigma} / \kappa)
    \end{align}
\end{subequations}
where only the sampling of the variance is i.i.d.\ since it enters via the scaling parameter $\kappa$ in the likelihood of the mean.
Using $\bm{\Sigma}_0$ rather than $\bm{\Psi}$ corresponds to parametrizing the distribution of $\bm{\Sigma}$ with an inverse $\chi^2$- rather than a $\Gamma^{-1}$-distribution in the univariate case, which has the advantage of a clearer interpretation of $\bm{\Sigma}_0$.

In Appendix~\ref{sec:appendix_niwshapeinter} we derive that taking a \texttt{NIG} (\texttt{NIW}) distribution as a conjugated prior corresponds to assuming prior knowledge about the mean and the variance extracted from $\kappa$ virtual measurements of the former and $2\alpha$ virtual measurements ($\nu$ virtual measurements) for the latter.
Therefore, it appears natural to define the total evidence of the prior as the sum of the number of virtual measurements:
\begin{align}
    \Phi' &:= \kappa + 2\alpha &
    \Phi' &:= \kappa + \nu
\end{align}
where the former (latter) refers to the univariate (multivariate) case.
In~\cite{amini20} the authors define the total evidence of the univariate case as\footnote{In the notation of~\cite{amini20} $\kappa$ becomes $\nu$ and the total evidence reads $\Phi = 2\nu + \alpha$.}
\begin{equation}
    \Phi := 2 \kappa + \alpha \,.
\end{equation}
For consistency we therefore propose to change this and to adopt our definition of $\Phi'$.
We will revisit this definition in the next section and discuss it in the context of the evidence regularizer.

\subsection{Ambiguity of shape parameters}
We follow the approach in~\cite{amini20} and use the posterior predictive or model evidence of a \texttt{NIW} distribution for finding a proper loss function $\mathcal{L}_i^\text{NLL}$.
From Bayesian probability theory the model evidence is a marginal likelihood and, as such, defined as the likelihood of an observation, $\vec{y}_i \in \mathbb{R}^n$, given the evidential distribution parameters, $\mathfrak{m}=(\vec{\mu}_0, \bm{\Psi}, \kappa, \nu)$, and is computed by marginalizing over the likelihood parameter (a.k.a. the nuisance parameter), $\bm{\theta} = (\vec{\mu}, \bm{\Sigma})$, where $\kappa, \nu \in \mathbb{R}$, $\vec{\mu}, \vec{\mu}_0 \in \mathbb{R}^n$, and $\bm{\Sigma}, \bm{\Psi}$ are positive definite $\mathbb{R}^{n \times n}$ matrices.
In our case of placing a \texttt{NIW} evidential prior on a multivariate Gaussian likelihood function an analytical solution exists and can be parametrized with a multivariate $t$-distribution with $\nu - n + 1$ DoF (see Appendix~\ref{sec:appendix_modev} and \ref{sec:appendix_derivmodev} for more details):
\begin{equation}
    \label{eq:modev}
    p(\vec{y}_i | \mathfrak{m})
    = \int \! \mathrm{d} \bm{\theta} \, \mathcal{N}(\vec{y}_i | \bm{\theta}) \,  \operatorname{\texttt{NIW}}(\bm{\theta} | \mathfrak{m}) 
    = t_{\nu-n+1} \left( \vec{y}_i \middle| \vec{\mu}_0, \frac{1}{\nu - n + 1} \frac{1 + \kappa}{\kappa} \bm{\Psi} \right).
\end{equation}
Using this result we can compute the negative log-likelihood loss $\mathcal{L}^{\text{NLL}}_i$ for sample $i$ as:
\begin{align}
    \mathcal{L}^{\texttt{NLL}}_i = - \log p(\vec{y}_i | \mathfrak{m}) 
    &= \log \Gamma\!\left( \frac{\nu - n + 1}{2} \right) - \log \Gamma\!\left( \frac{\nu + 1}{2} \right) \nonumber \\
    &\phantom{=} + \frac{n}{2} \log\!\left( \pi \, \frac{1 + \kappa}{\kappa} \right) - \frac{\nu}{2} \log |\bm{\Psi}| \nonumber \\
    &\phantom{=} + \frac{\nu + 1}{2} \log \left| \bm{\Psi} + \frac{\kappa}{1 + \kappa} (\vec{y}_i - \vec{\mu}_0) (\vec{y}_i - \vec{\mu}_0)^\top \right| \,. \label{eq:nll}
\end{align}

From the compact notation in Eq.~\eqref{eq:modev} it is obvious that $p(\vec{y}_i | \mathfrak{m})$ on its own is not capable of defining $\mathfrak{m}$ unambiguously.
In particular, a fitting approach could be used to find $\nu$, $\vec{\mu}_0$ and the product $(1+\kappa)/\kappa \, \bm{\Psi}$ from data.
However, in order to disentangle the latter additional constraints have to be set, e.g., via an additional regularization of $\kappa$.

The higher-order evidential distribution is projected by integrating out the nuisance parameters $\vec{\mu}$ and $\bm{\Sigma}$ and, in the univariate case, the four DoF of $\mathfrak{m}$ collapse into three DoF of a scaled Student's $t$-distribution.
Fitting this reduced set of DoF is not sufficient to recover all DoF of the evidential distribution.
The impact of this observation is that fitting the width of the $t$-distribution will not help to unfold $\kappa$ and $\bm{\Psi}$ and it is possible to find manifolds with different values of $\kappa$ and $\beta$ but with the same value for the loss function $\mathcal{L}_i^\text{NLL}$.
In fact, $\kappa$ can be tuned such that for any given value of $\mathcal{L}_i^\text{NLL}$ a value for $\bm{\Psi}$ or $\beta$ can be find.
Therefore, $\mathcal{L}_i^\text{NLL}$ on its own is not sufficient to learn the parameters $\mathfrak{m}$.
(See Appendix~\ref{sec:appendix_degeneration} for more details.)

In~\cite{amini20} this degeneration of the loss function is broken by introducing the evidence regularizer,
\begin{equation}
    \mathcal{L}_i^\text{R} = |y_i - \mu_0| \Phi \,,
\end{equation}
however, it is unclear how the NN could learn the parameters $\mathfrak{m}$ when the evidence regularizer is disabled by setting $\lambda = 0$.
More importantly, although $\mathcal{L}_i^\text{R}$ breaks the degeneration, using $\Phi = 2\kappa + \alpha$, the total loss $\mathcal{L}_i = \mathcal{L}_i^\text{NLL} + \lambda \mathcal{L}_i^\text{R}$ can easily be minimized (in theory) for $\kappa$ by simply sending $\kappa \to 0$ since any impact on $\mathcal{L}_i^\text{NLL}$ can be compensated by adjusting $\beta$ without changing the value of $\mathcal{L}_i^\text{NLL}$.
Sending $\kappa$ to zero drives the ratio of aleatoric and epistemic uncertainty to zero as well, cf. Eqs.~\eqref{eq:uncs}, thus making their values useless.
(In practice, $\kappa \to 0$ is numerically unstable and minimizer we fail to converge towards this point.)

In summary: The loss $\mathcal{L}_i^\text{NLL}$ is degenerated and requires regularization of either $\kappa$ or $\bm{\Psi}$.
The proposed regularizer of~\cite{amini20} does not lead to correct uncertainty estimations and has a numerical unstable minimum.

\subsection{Challenging extraction of shape parameters}
\begin{figure}[htbp]
    \centering
    \begin{subfigure}{.45\textwidth}
        \centering
        \includegraphics[width=\textwidth]{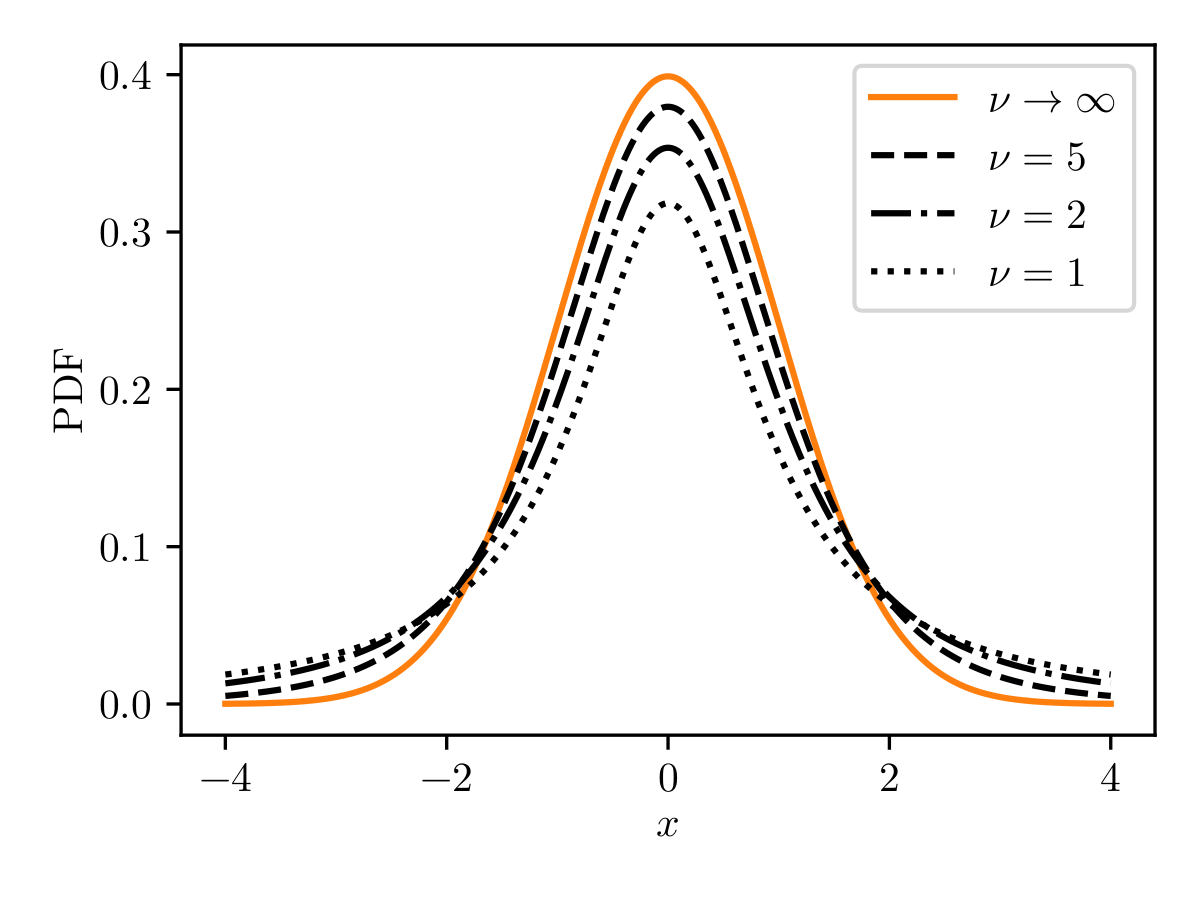}
        \caption{Student's $t$-distributions.}
        \label{fig:students}
    \end{subfigure}
    \begin{subfigure}{.45\textwidth}
        \centering
        \includegraphics[width=\textwidth]{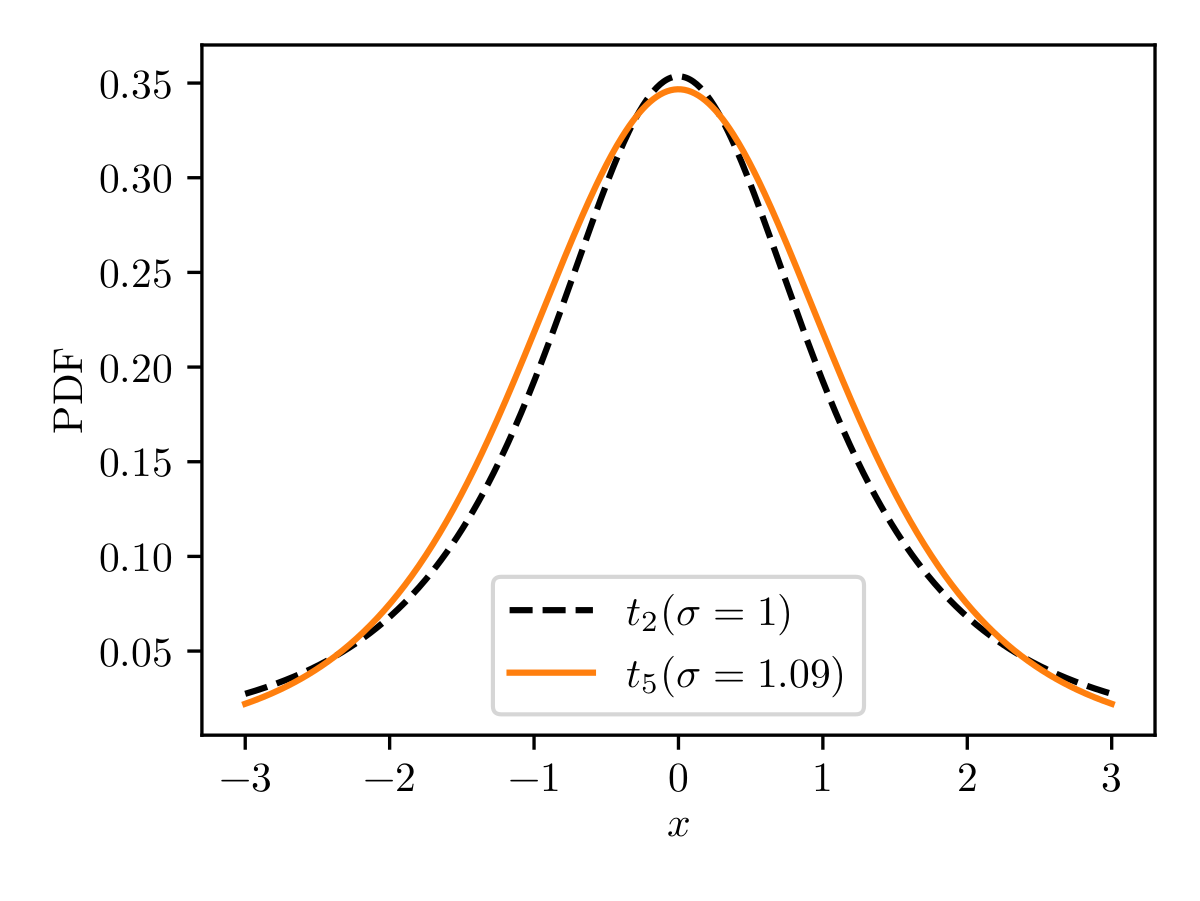}
        \caption{Fit of a $t_2$ distribution using a $t_5$ function.}
        \label{fig:dfit}
    \end{subfigure}
    \caption{(Left) Student's $t$-distribution for different values of $\nu$ and (right) fit of a $t$-distribution with $\nu=5$ to a $t$-distribution with $\nu=2$.}
\end{figure}
We argued previously that minimizing Eq.~\eqref{eq:nll} is nothing but the fit of a $t$-distribution to data.
In Appendix~\ref{sec:gfit} we further elaborate that technically only a single data point is fitted per distribution, although correlations between neighboring points will enter in practice.
It is therefore difficult to estimate the number of data points used per fit, still, we argue here that a large statistic is necessary to extract the parameter $\nu$ reliably which plays a crucial role in estimating the epistemic uncertainty.

In general, small values for $\nu$ will raise the tails of the distribution but only slightly affect the shape of the core of the function where most measurements will be find, assuming that these indeed follow a $t$-distribution.
Extraction of $\nu$ using a fit thus needs a large statistic which properly describes the tails.
In reality, however, the assumption of normal distributed events often collapses especially in the tails of a distribution which makes a fit of $\nu$ even more ambitious.
Assuming one only has a decent statistic near the core, the parameters $\nu$ and $\sigma$ of a scaled Student's $t$-distribution,\footnote{For the sake of brevity we overload here the notations for $\mu$ and $\sigma$.} $\operatorname{St}_\nu(y_i | \mu, \sigma)$, are highly correlated as shown in Fig.~\ref{fig:dfit} where we fit $\sigma$ of a $t$-distribution with $\nu=5$ to a $t$-distribution with $\nu=2$ on the interval $x \in [-3, 3]$.

To study possible biases of the fitted values of the parameters $\nu$ and $\kappa$ we conduct a pseudo experiment where we generate data by drawing them i.i.d.\ from Student's $t$-distributions with fixed shape parameters and fit them on different sample sizes.
(See Appendix \ref{sec:appendix_nubias} for details.)
For each sample size we evaluate 200 fits and find a bias for $\nu$ and $\sigma$ for small sample sizes which decreases if the sample size increases.

In summary: Extracting the parameter $\nu$ of a $t$-distribution requires a sufficiently large data sample.
Due to correlations the fit result is biased which can be significant if the sample size is too small.
We showed this for the univariate case (where $\nu$ corresponds to $2\alpha$) but the same holds for the multivariate case as well where an even larger data set is needed.

\section{A possible solution}
\label{sec:solution}
In this section we describe a possible solution for two of the aforementioned issues.
The issue regarding the correlation of the shape parameters of a $t$-distribution is not affected by our solution proposal and biases have to be studied using data.

In~\cite{amini20} the authors did not introduce the term $\mathcal{L}_i^\text{R}$ as a way to lift the degeneration of $\mathcal{L}_i^\text{NLL}$ but motivate it as an \textit{evidential regularizer}, similar to~\cite{sensoy18}.
The idea of combining the $\ell_1$-norm of the prediction error with the total evidence $\Phi$ is to enforce the NN to learn large errors in the prediction are acceptable, as long as this is reflected in a small total evidence and vice-versa.
In other words, in the absence of data which would have the potential to squeeze the prediction error, the prior information should approach an uninformed prior and $\mathcal{L}_i^\text{R}$, as proposed by the authors, is one possible metric to measure the distance to it but does not lead to a meaningful minimum as described previously.
Finding a suitable metric which breaks the degeneration of $\mathcal{L}_i^\text{NLL}$ but also pushes the distribution towards an uninformed prior is not straight-forward.
For example, instances of the $f$-divergence family, differential entropy or taking the peak height of the function as a measure to meet the second requirement do not break the degeneration.
This is because any metric of $t_\nu(\vec{\mu}_t, \bm{\Sigma}_t)$ is, by construction, ignorant of internal dependencies of the shape parameters but in order to break the degeneration $\kappa$ and $\bm{\Sigma}_0$ have to be unfolded from $\bm{\Sigma}_t = \bm{\Sigma}_t(\nu, \kappa, \bm{\Sigma}_0)$.

We therefore propose to acknowledge the loss of one DoF and couple the parameters $\kappa$ and $\nu$ with a constant hyperparmeter $r$, i.e., using again the index notation:
\begin{equation}
    \label{eq:kappanucoupling}
    \nu_i = r \kappa_i \,.
\end{equation}
Including an evidential regularizer as $|\vec{y}_i - \vec{\mu}_{0,i}| \nu_i$ to the loss function is therefore no longer necessary and minimizing $\mathcal{L}_i = \mathcal{L}_i^\texttt{NLL}$ is sufficient.
We motivate this ansatz by considering it unnatural to have prior information from $\kappa$ virtual measurements for the mean and $\nu$ virtual measurements for the variance where the ratio $\kappa / \nu$ significantly fluctuates throughout the data sample or even differs largely in scale.
Another way of seeing the implicit coupling of both variables is the case of vanishing epistemic uncertainty, i.e., $\kappa \to \infty$.
The model should then become a normal distribution as described in Appendix~\ref{sec:gfit} which corresponds to $\nu \to \infty$.
(This is $\alpha \to \infty$ in the univariate case.)
Coupling $\kappa$ and $\nu$, as proposed in Eq.~\eqref{eq:kappanucoupling}, enforces this behavior.

In summary: For the univariate case we propose to couple the parameters $\kappa$ and $\alpha$ via a hyperparameter $r$ that is kept constant for all instances.
Similarly, in the multivariate case one should couple $\kappa$ and $\nu$ with $r$.
In the next section we summarize how this changes the loss function.

\section{Multivariate generalization}
\label{sec:multi}
In this section we summarize our multivariate generalization, combine it with our proposed solution from Sec.~\ref{sec:solution} and benchmark it with a multivariate experiment.
In general, using our proposed multivariate generalization it is possible to not just learn uncertainties of each regression target of a multivariate dataset individually, but also to learn their correlations.
Whereas using chained univariate regressions it is possible by sampling to extract the correlation of $(\vec{y}_i, \vec{y}_j) \in \mathbb{R}^{n \times n}$ pairs at $(x_i, x_j)$ with $i \neq j$, it is impossible by the very nature of the univariate distributions to get the correlation of $(y_{ij}, y_{ik}) \in \mathbb{R}^{1 \times 1}$ at $(x_i, x_i)$.
The latter, i.e., the feature correlation at each point $x_i$, can only be extracted with a multivariate approach.

Taking a \texttt{NIW} distribution as the conjugated prior we found that minimizing the loss
\begin{align}
    \mathcal{L}_i \equiv \mathcal{L}^{\texttt{NLL}}_i
    &= \log \Gamma\!\left( \frac{\nu_i - n + 1}{2} \right) - \log \Gamma\!\left( \frac{\nu_i + 1}{2} \right) \nonumber \\
    &\phantom{=} + \frac{n}{2} \log(r + \nu_i) - \nu_i \sum\limits_j \ell_j^{(i)} \nonumber \\
    &\phantom{=} + \frac{\nu_i + 1}{2} \log \! \Big| \bm{L}_i \bm{L}_i^\top + \frac{1}{r + \nu_i} (\vec{y}_i - \vec{\mu}_{0,i}) (\vec{y}_i - \vec{\mu}_{0,i})^\top \Big| + \text{const.}
\end{align}
allows the estimation of the prediction and both types of uncertainties as:\footnote{See Appendix~\ref{sec:moments} for a derivation.}
\begin{equation}
    \underbrace{\mathbb{E}[\mu] = \vec{\mu}_{0,i}}_{\text{prediction}}
    \qquad\qquad
    \underbrace{\mathbb{E}[\bm{\Sigma}] \propto \frac{\nu_i}{\nu_i - n - 1} \bm{L}_i \bm{L}_i^\top}_{\text{aleatoric}}
    \qquad\qquad
    \underbrace{\operatorname{var}[\vec{\mu}] \propto \mathbb{E}[\bm{\Sigma}] / \nu_i}_{\text{epistemic}}
\end{equation}
where we rewrote the positive (semi-)definite matrix $\bm{\Psi}_i = \nu_i \bm{\Sigma}_{0,i} = \nu \bm{L}_i \bm{L}_i^\top \in \mathbb{R}^{n \times n}$ with the lower triangular matrix $\bm{L}_i$ and enforce positive diagonal elements by parametrizing with $\vec{\ell}^{\,(i)} \in \mathbb{R}^{n(n+1)/2}$:
\begin{equation}
    \left( \bm{L}_i \right)_{jk} =
    \begin{cases}
        \exp \! \left\{ \ell_j^{(i)} \right\} & \text{if } j=k, \\
        \ell_{jk}^{(i)} & \text{if } j>k, \\
        0 & \text{else.}
    \end{cases}
\end{equation}
We note that regardless of this rewriting the limit $\nu_i \to \infty$ is numerically unstable and cut-offs have to be placed in practice.

In order to learn the parameters $\mathfrak{m}_i = (\vec{\mu}_{0,i}, \vec{\ell}^{\,(i)}, \nu_i)$ a NN has to have $n(n + 3)/2 + 1$ output neurons and one hyperparameters $r$.
By using Eq.~\eqref{eq:kappanucoupling} we acknowledge the loss of one DoF due to the projection of the higher-order \texttt{NIW} distribution.
This reduction comes with the cost that we loose the predictive power on a global scale of the aleatoric and the epistemic uncertainties which we assume to be constant by taking $r$ as a hyperparameter.
In practice this means that if a global scale is of interest, one has to rescale the predicted uncertainties (which is viable with either Bayesian or Frequentist techniques).

\begin{figure}[htbp]
    \centering
    \begin{subfigure}{.49\textwidth}
        \centering
        \includegraphics[scale=.8]{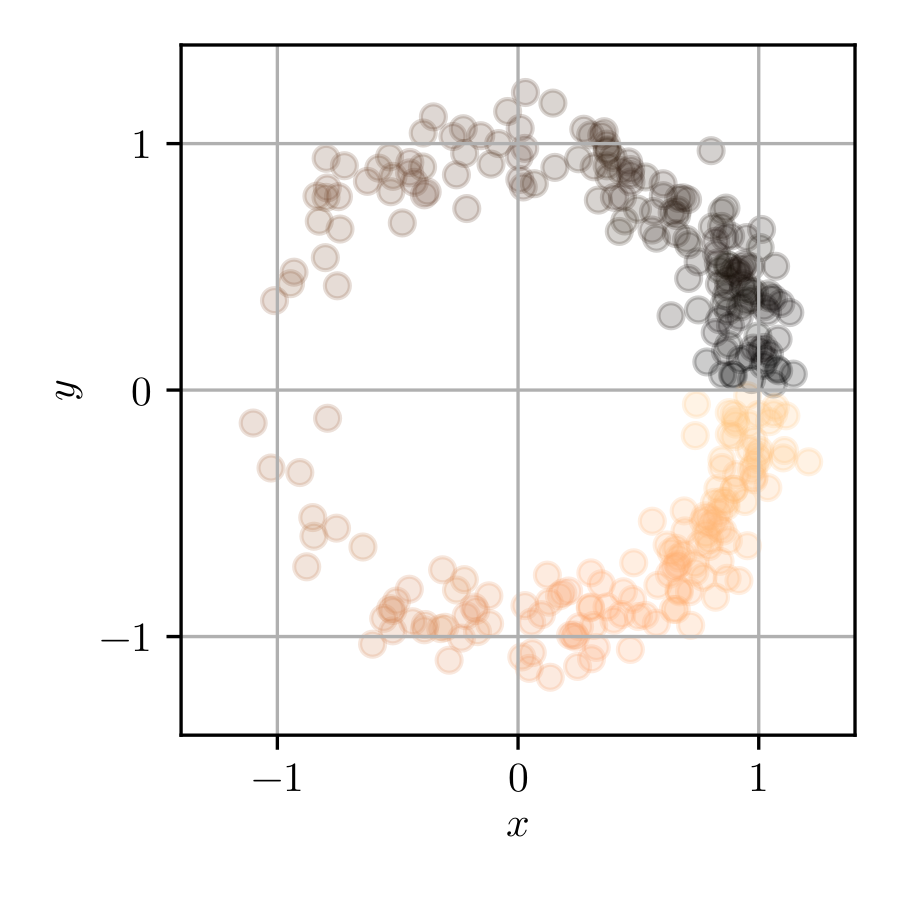}
        \caption{Synthetic data}
        \label{fig:data}
    \end{subfigure}
    \begin{subfigure}{.49\textwidth}
        \centering
        \includegraphics[scale=.8]{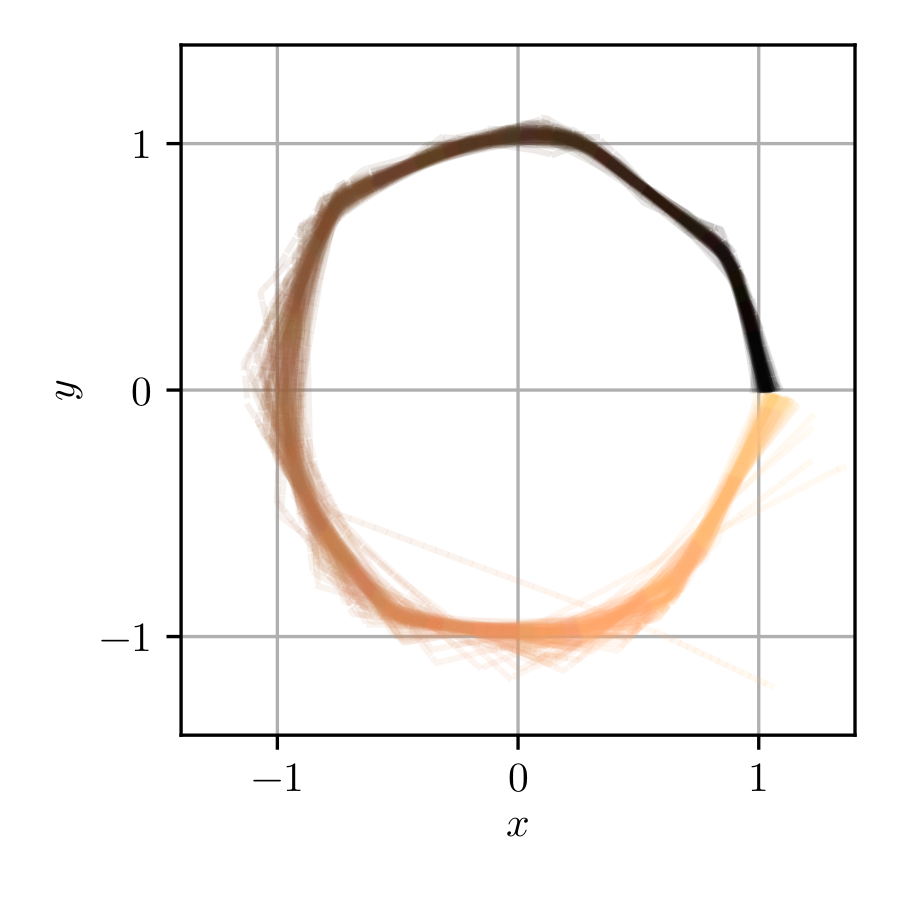}
        \caption{Predictions of NNs}
        \label{fig:predictions}
    \end{subfigure}
    \caption{(Left) The distribution of our test data in the $xy$-plane. The value of $t$ is color coded (see Fig.~\ref{fig:data_xyt} for a 3d representation). (Right) the overlayed prediction of 100 trained NNs for given values of $t$.}
\end{figure}

Finally, we conduct a simple experiment with $n=2$ and $r=1$ to benchmark our multivariate generalization.
Our experiment is built upon the univariate experiment described by Amini et al.~\cite{amini20} with a critical difference:
Rather than training and evaluating the NN on partially disjunct data samples\footnote{In~\cite{amini20} the NN is trained for $t \in [-4, +4]$ but evaluated on $t \in [-7, +7]$.} where the NN has no chance to identify an increasing epistemic uncertainty, we generate data in the $xy$-plane with varying density.
We overload our notation of $x$ and $y$ now being the features of our data sample given input $t$,
\begin{subequations}
\begin{align}
    x &= (1 + \epsilon) \cos t
    &
    y &= (1 + \epsilon) \sin t \,,
\end{align}
where the distribution of $t$ is not flat but has a $\vee$ shape,
\begin{align}
    t &\sim
    \begin{cases} 
        1 - \frac{\zeta}{\pi} & \text{if } \zeta \in [0, \pi] \,, \\
        \frac{\zeta}{\pi} - 1 & \text{if } \zeta \in (\pi, 2\pi] \,, \\
        0 & \text{else,}
    \end{cases}
\end{align}
\end{subequations}
with uniformly distributed $\zeta \in [0, 2\pi]$ and $\epsilon$ is drawn from a normal distribution, $\epsilon \sim \mathcal{N}(0, 0.1)$.
We draw 300 data points in total (see Fig.~\ref{fig:data}) and fit the distribution with a small, fully connected NN with a single input neuron, two hidden layers of 32 neurons using Rectified Linear Unit activation functions, and six output neurons.
The output of the last layer, $\vec{p} \in \mathbb{R}^6$, is transformed and subsequently taken as the parameters of the evidential distribution:
\begin{equation}
    \vec{\mu} = \begin{pmatrix} p_1 \\ p_2 \end{pmatrix},
    \quad
    \bm{L} = \begin{pmatrix} \exp \{ p_3 \} & 0 \\ p_4 & \exp \{ p_5 \} \end{pmatrix},
    \quad
    \nu = 8 + 5 \tanh p_6 \,,
\end{equation}
where an exponential function is used to constrain the diagonal elements of $\bm{L}$ to be strictly positive and the transformation of $\nu$ corresponds to the required lower bound\footnote{See Appendix~\ref{sec:moments} for details.} $\nu > n + 1 = 3$ and a cut-off $\nu < 13$ -- empirically, $t$-distributions with more than $13$ DoF are almost indistinguishable from genuine normal distributions.

Constraining $\nu$ onto the interval $\nu \in (3, 13)$ makes this parameter effectively a gate.
Being closed, $\nu \approx 3$, corresponds to the situation that the additional DoF of a $t$-distribution helps to better fit the data, whereas being open, $\nu \gg 3$, indicates that the genuine distribution function actually yields the better fit result.
That is, even though the data are drawn from a normal distribution, a fit with a more flexible function will more likely find a better minimum for sparsely sampled data.
This extra flexibility of the $t$-distribution w.r.t.\ a normal distribution, coming from the extra DoF, becomes less important when the data distribution becomes denser and, on average, better resembles its genuine distribution function.

\begin{figure}[htbp]
    \centering
    \includegraphics[scale=.8]{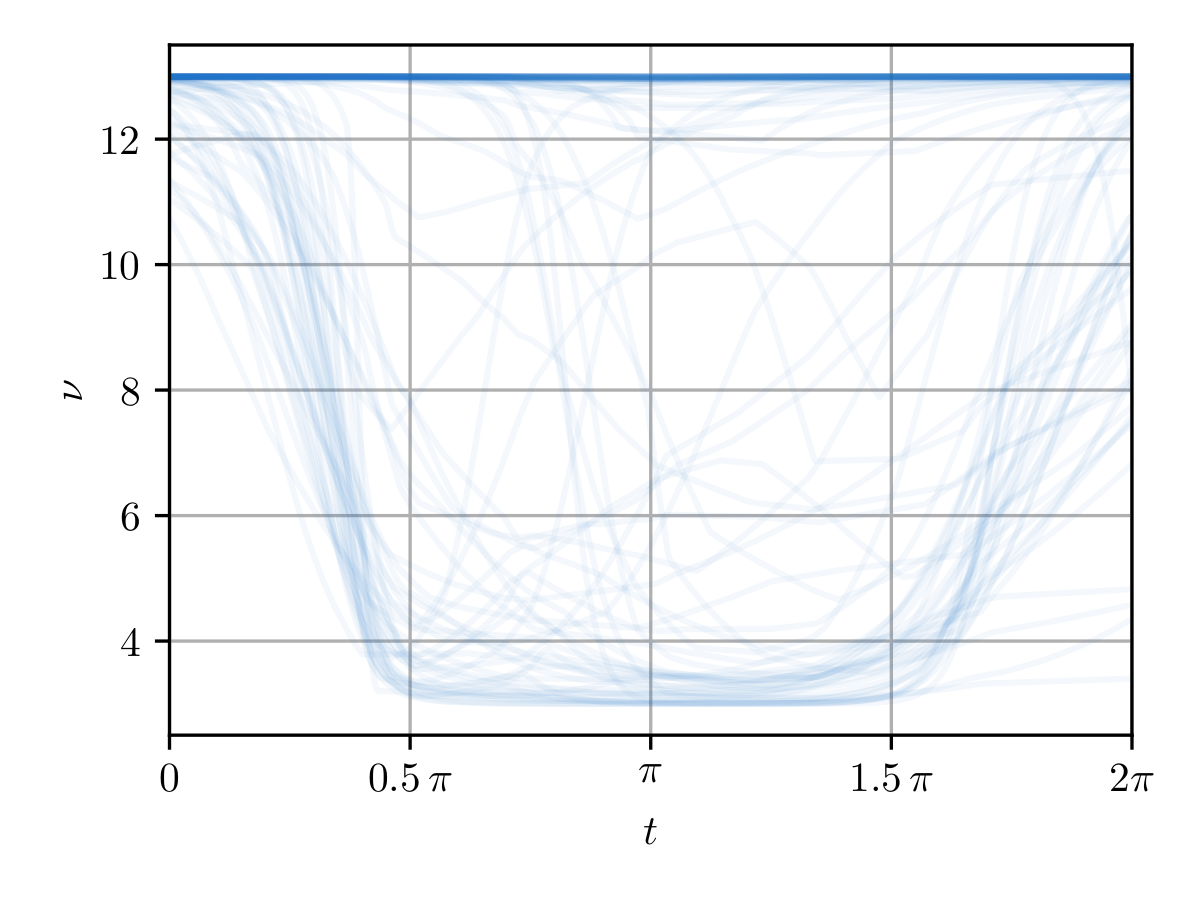}
    \caption{Overlayed parameter $\nu$ of 100 trained NNs. Low (high) values correspond to a large (low) epistemic uncertainty.}
    \label{fig:nu}
\end{figure}

In total we train 100 NNs from scratch on the synthetic data sample and overlay their predictions for $x = \mu_1$ and $y = \mu_2$ in Fig.~\ref{fig:predictions}.
More importantly for this work is the behavior of the parameter $\nu$ which we show in Fig.~\ref{fig:nu}.
We find a statistical significant drop in $\nu$ towards the center of $t$ which corresponds to an enhancement of the epistemic uncertainty.
It is exactly this area where the data distribution becomes sparse and therefore nicely meets our expectations.
However, not all NNs converged to this solution and fluctuations are present.
In fact, we find the shown behavior being strongly correlated with the sample size of the synthetic data sample: reducing the sample size causes serious over-fitting of the model, whereas increasing quickly opens the $\nu$ gate for all values of $t$.
We find that in presence of a sufficient amount of data not just the $\nu$ gate is open for all values of $t$, but the predicted values of $x$ and $y$ resemble the genuine distribution well, also in the regions $0.5\pi \lesssim t \lesssim 1.5\pi$ (we already see this behavior indicated in Fig.~\ref{fig:predictions}) which is again in agreement with our expectations.

The outlined technique therefore helps to detect variations of the epistemic uncertainty throughout the data landscape.
See Appendix~\ref{sec:appendix_exp} for more details.

\section{Related Work}
\label{sec:rw}
Our work builds specifically on the prior art~\cite{amini20} for uncertainty estimation with evidential neural networks, and more generally on the advancing area of uncertainty reasoning in deep learning.

The probabilistic perspective in machine learning (ML) frames learning as inferring plausible models to explain observed data.
Observed data can be consistent with many models, and therefore which model is appropriate given the data is uncertain~\cite{ghahramani15}.
Probabilistic (or Bayesian) ML methods are rooted in probability theory and thus provide a framework for modeling uncertainties.
Traditional probabilistic methods include Gaussian processes~\cite{rasmussen06}, latent variable models~\cite{bishop98}, and probabilistic graphical models~\cite{koller09}.
In recent years there have been many explorations into Bayesian approaches to deep learning~\cite{kendall17,neal96,guo17,wilson15,hafner18,ovadia19,izmailov19,seedat19}.
The key observation is that neural networks are typically underspecified by the data, thus different settings of the parameters correspond to a diverse variety of compelling explanations for the data -- i.e., a deep learning posterior consists of high performing models which make meaningfully different predictions on test data, as demonstrated in~\cite{izmailov19,garipov18,zolna19}.
This underspecification by NNs makes Bayesian inference, and by corollary uncertainty estimation, particularly compelling for deep learning.
Bayesian deep learning aims to compute a distribution over the model parameters during training in order to quantify uncertainties,
such that the posterior is available for uncertainty estimation and model calibration~\cite{guo17}.
With Bayesian NNs that have thousands and millions of parameters this posterior is intractable, so implementations largely focus on several approximate methods for Bayesian inference:
First, Markov Chain Monte Carlo (MCMC) methods iteratively draw samples from the unknown posterior distribution, and efficient MCMC methods make use of gradient information rather than performing random walks.
In particular stochastic gradient MCMC for Bayesian NNs~\cite{welling11,li16,park18,maddox19}, with a main drawback being the inability to capture complex distributions in the parameter space without increasing the computational overhead.
Second, variational inference (VI) performs Bayesian inference by using a computationally tractable \textit{variational} distribution to approximate the posterior.
One approach by Graves et al.~\cite{graves13} is to use a Gaussian variational posterior to approximate the distribution of the weights in a network, but the capacity of the uncertainty representation is limited by the variational distribution.
In general we see that MCMC has a higher variance and lower bias in the estimate, while VI has a higher bias but lower variance~\cite{mattei19}.
The preeminent Bayesian deep learning approach by Gal and Ghahramani~\cite{gal16} showed that variational inference can be approximated without modifying the network.
This is achieved through a method of approximate variational inference called Monte Carlo Dropout (MCD), whereby dropout is performed during inference, using multiple dropout masks.

Better understanding of the integration of deep learning with probabilistic ML such as Gaussian processes (GP) is also a fruitful direction, namely with Deep Kernel Learning~\cite{wilson15,lavin21} and deep GP~\cite{duvenaud14,dutordoir21}.

Alternative to the prior-over-weights approach of Bayesian NN, one can view deep learning as an evidence acquisition process -- different from the Bayesian modeling nomenclature, evidence here is a measure of the amount of support collected from data in favor of a sample to be classified into a certain class, and uncertainty is inversely proportional to the total evidence~\cite{sensoy18}.
Samples during training each add support to a learned higher-order, evidential distribution, which yields epistemic and aleatoric uncertainties without the need for sampling.
Several recent works develop this approach to deep learning and uncertainty estimation which put this in practice with \textit{prior networks} that place priors directly over the likelihood function~\cite{amini20,malinin18}.
These approaches largely struggle with regularization~\cite{sensoy18}, generalization (particularly without using out-of-distribution training data)~\cite{malinin18,hafner18}, capturing aleatoric uncertainty~\cite{gurevich19}, and the issues we have addressed above with the prior art Deep Evidential Regression~\cite{amini20}.

There are also the frequentist approaches of bootstrapping and ensembling, which can be used to estimate NN uncertainty without the Bayesian computational overhead as well as being easily parallelizable -- for instance Deep Ensembles, where multiple randomly initialized NNs are trained and at test time the output variance from the ensemble of models is used as an estimate of uncertainty~\cite{lakshminarayanan17}.

\section{Conclusion}
We discussed the recent developments towards uncertainty-aware neural networks, \textit{Deep Evidential Regression}~\cite{amini20}, identifying several outstanding issues with the approach and providing detailed solutions grounded in theory and experimental results.
In addition to correcting the prior art, we extend it for multivariate scenarios.
The solutions and new approach we presented here would benefit from future studies towards empirical validation.

\section*{Broader Impact}
Neural networks have already had significant impacts in many applications -- from medical imaging to dialogue systems to autonomous vehicles -- and will continue to do so for years to come.
Yet there are important shortcomings in our understanding and confidence in this class of machine learning, notably in the ability to estimate uncertainties and calibrate models.
This problem becomes more complex, and potentially dangerous, when NNs are built within larger systems that combine data, software, hardware, and people in dynamic, complex ways.
Reliable and systematic methods of uncertainty quantification with NNs are needed, especially considering deployments in safety critical domains such as medicine and autonomous vehicles.
In addition to the practical utility of reliably quantifying uncertainty in NNs, there is a significant need to build confidence in the models and establish trust with the end-users.
Work towards quantifying uncertainties in machine learning is essential such that these models and systems ``know when they don't know'', and are thus more trusted and usable in real-world scenarios.

\section*{Reproducibility}
The source code for reproducing our experiments -- implementation of the NN and multivariate methods, and algorithms to generate the data -- is available on \href{https://github.com/avitase/mder/}{\texttt{github.com/avitase/mder/}} (MIT License).
The model and experiments are lightweight, running locally on a 4-core MacBook Pro in under an hour.
We use the \href{https://www.python.org/}{\texttt{Python}} programming language as well as various libraries, most notably \href{https://pytorch.org/}{\texttt{PyTorch}}, \href{https://numpy.org/}{\texttt{NumPy}}, \href{https://jupyter.org/}{\texttt{Jupyter}} and \href{https://matplotlib.org/}{\texttt{matplotlib}}.

\newpage
\appendix
\section{Maximum likelihood estimation using Gaussians}
\label{sec:gfit}
A simple approach to estimate uncertainties of a regression-based NN is to assume that data are drawn i.i.d.\ from normal distributions, i.e.,
\begin{equation}
    y_i \sim \mathcal{N} \big( \mu_i, \sigma_i^2 \big) \equiv \mathcal{N} \big( \mu(x_i), \sigma^2(x_i) \big)
\end{equation}
for each $(x_i, y_i)$ pair of the data sample at hand.
Instead of using Bayesian inference one could simple seek for the maximum of the combined likelihood
\begin{equation}
    \max\limits_{\bm{w}} L(\bm{w}) = \prod\limits_i \mathcal{N} \big( y_i \big| \mu_i, \sigma_i^2 \big) \,,
\end{equation}
or, alternatively, for the minimum of the negative log-likelihood
\begin{equation}
    \label{eq:minnllgaussian}
    \min\limits_{\bm{w}} \mathcal{L}(\bm{w}) = \sum\limits_i \underbrace{\left( \frac{1}{2} \log (2\pi\sigma_i^2) + \frac{(y_i - \mu_i)^2}{2 \sigma_i^2} \right)}_{\mathcal{L}_i(\bm{w})}
\end{equation}
and let the NN itself estimate $\mu_i$ and $\sigma_i^2$ by adding one extra neuron to the output layer and interpret the output values of these two neurons as $\mu_i \equiv \mu(x_i, \bm{w})$ and $\sigma_i^2 \equiv \sigma^2(x_i, \bm{w})$.

\begin{figure}[htbp]
    \centering
    \begin{subfigure}{.45\textwidth}
        \centering
        \includegraphics[width=\textwidth]{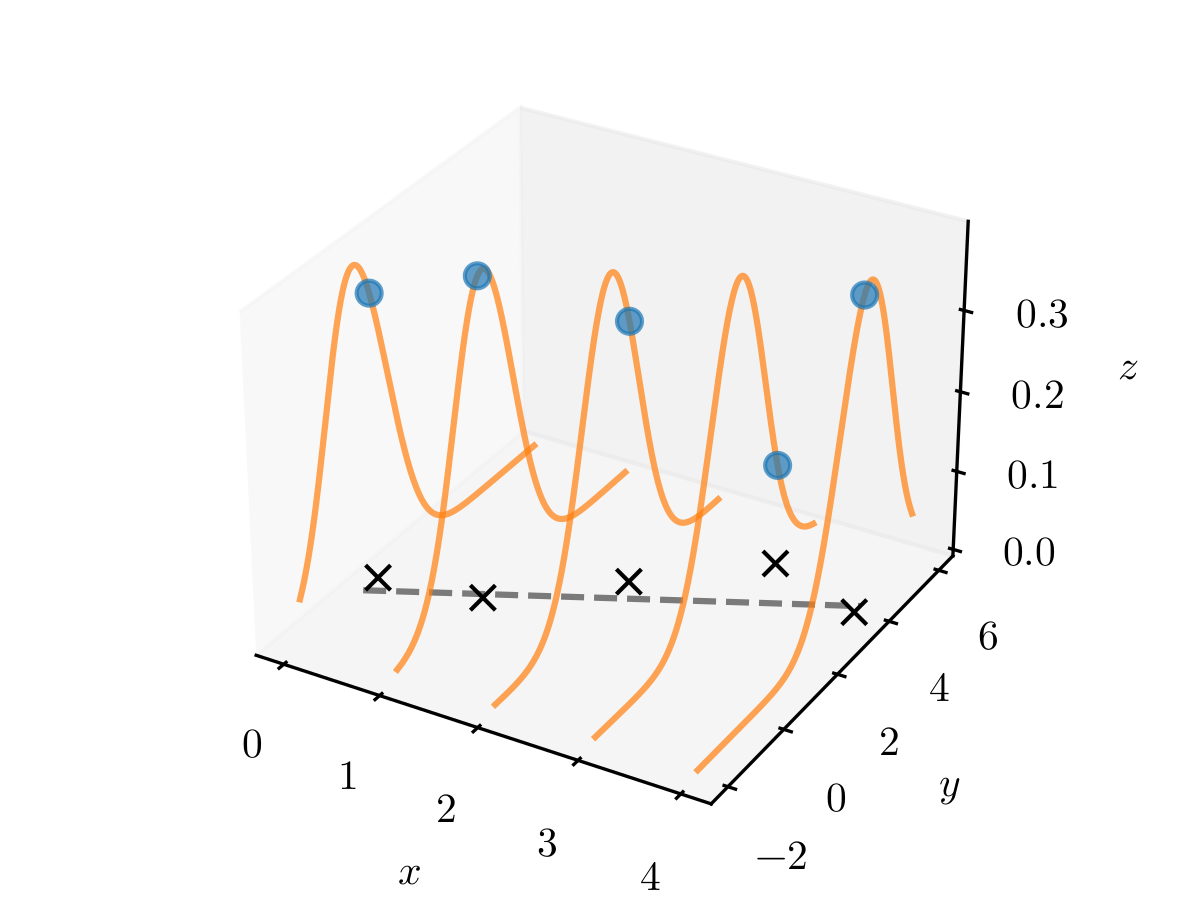}
    \end{subfigure}
    \begin{subfigure}{.45\textwidth}
        \centering
        \includegraphics[width=\textwidth]{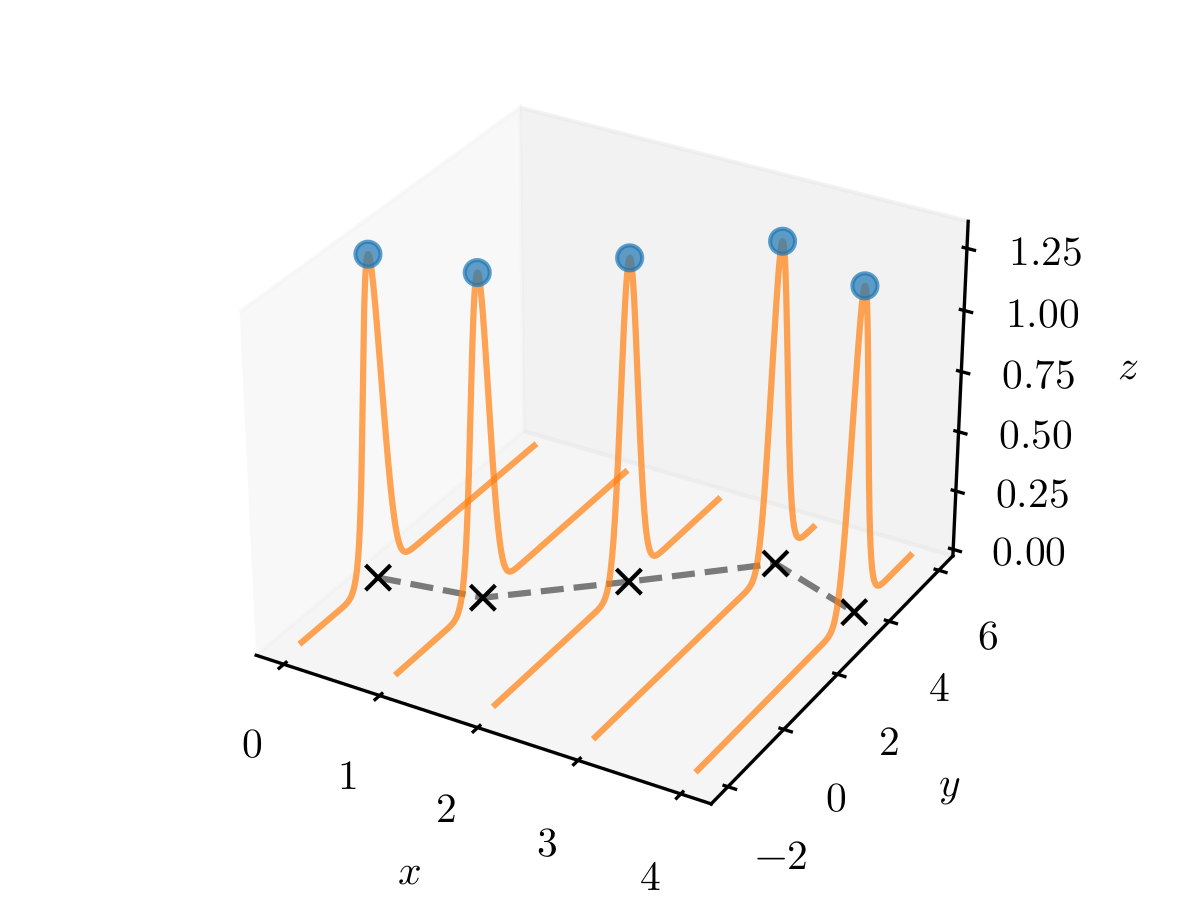}
    \end{subfigure}
    \caption{Fitting a data distribution in $x$ and $y$ using Eq.~\eqref{eq:minnllgaussian}. The prediction of the fitted model is shown as a dashed line in the $x$-$y$ plane and the value of $\mathcal{L}_i$ as a solid line in the $y$-$z$ plane. The fit on the right over-fits the data but has the smaller total loss $\mathcal{L}$.}
    \label{fig:greg}
\end{figure}

We note that this corresponds to fitting Gaussian functions to single data points and it is the objective of the supervisor of the training process of the NN to ensure that $\mu(x_i, \bm{w})$ and $\sigma^2(x_i, \bm{w})$ do not over-fit the data as shown in the right side of Fig.~\ref{fig:greg}.
Here, the model is obviously unable to differentiate between aleatoric and epsitemic uncertainty and will merge both components into $\sigma_i$ if the model is under-fitting.
In case of missing data the model will likely interpolate between regions where data are available and thus underestimate the epistemic uncertainty.

For the sake of brevity we drop the index notation of the parameters of the likelihood in the main text but stress the tight coupling of each $(x_i, y_i)$ pair to its individual $(\mu, \sigma^2)$ pair.
In reality however, given a sufficiently large data sample, a well behaving NN will likely produce similar values for $(\mu, \sigma^2)$ if two pairs $(x_i, y_i)$ and $(x_j, y_j)$ are close.
Technically, this approach is equivalent to fitting independent functions to each data point but the NN will correlate adjacent points.

\newpage
\section{Interpretation of the shape parameters of \texttt{NIG} and \texttt{NIW} distributions}
\label{sec:appendix_niwshapeinter}
An interpretation of the shape parameters of a \texttt{NIG} or \texttt{NIW} distribution can be find by analyzing the joint posterior density after taking $m$ measurements, 
\begin{equation}
    \vec{y} \in \mathbb{R}^m
    \qquad\qquad
    \text{or}
    \qquad\qquad
    \bm{Y} = (y_{ij}) =
    \begin{pmatrix}
        \vec{y}_1^{\,T} \\
        \vdots \\
        \vec{y}_n^{\,T}
    \end{pmatrix}  \in \mathbb{R}^{n \times m} \,,
\end{equation}
i.e., multiplying the prior density by the normal likelihood yields the posterior density
\begin{align}
    p(\mu, \sigma^2 | \vec{y}\,) &= \texttt{NIG}(\mu'_0, \kappa';\, \alpha', \beta') &
    p(\vec{\mu}, \bm{\Sigma} | \bm{Y}) &= \texttt{NIW}(\vec{\mu}'_0, \kappa';\, \bm{\Psi}', \nu')
\end{align}
with
\begin{subequations}
    \label{eq:postdens}
    \begin{align}
        \kappa' &= \kappa + m &
        \kappa' &= \kappa + m \\
        2\alpha' &= 2\alpha + m &
        \nu' &= \nu + m \\
        \mu_0' &= \frac{1}{\kappa + m} \begin{pmatrix} \kappa \\ m \end{pmatrix} \begin{pmatrix} \mu_0 \\ \langle \vec{y} \, \rangle \end{pmatrix} &
        \vec{\mu}_0' &= \frac{1}{\kappa + m} \begin{pmatrix} \kappa \\ m \end{pmatrix} \begin{pmatrix} \vec{\mu}_0 \\ \langle \bm{Y} \, \rangle \end{pmatrix} \label{eq:postdens_mu} \\
        2\beta' &= 2\beta + \tilde{s} + m \frac{\kappa}{\kappa'} \, (\mu_0 - \langle \vec{y} \, \rangle)^2 &
        \bm{\Psi}' &= \bm{\Psi} + \bm{\tilde{S}} + m \frac{\kappa}{\kappa'} \, (\vec{\mu}_0 - \langle \bm{Y} \rangle) (\vec{\mu}_0 - \langle \bm{Y} \rangle)^\top \label{eq:postdens_var}
    \end{align}
\end{subequations}
where we introduced the squared sum of residuals (up to a scaling constant $(n-1)$ these are estimators of the sample variance)
\begin{align}
    \tilde{s} &= \sum\limits_{i=1}^m (y_i - \langle \vec{y}\, \rangle)^2 &
    \bm{\tilde{S}} &= \sum\limits_{i=1}^m \left( \vec{y}_i - \langle \bm{Y} \rangle \right) \left( \vec{y}_i - \langle \bm{Y} \rangle \right)^\top
\end{align}
and the expectation value
\begin{align}
    \langle \vec{y} \, \rangle &= \frac{1}{n} \sum\limits_{i}^n y_i \in \mathbb{R} &
    \langle \bm{Y} \rangle &= \frac{1}{n} \sum\limits_{i,j=1}^{m,n} y_{ij} \, \vec{\mathrm{e}}_i \in \mathbb{R}^m \,.
\end{align}
These relations can easily be interpreted as the combination of prior information and the information contained in the data.
In particular, Eq.~\eqref{eq:postdens_mu} reads as the weighted sum of two measurement outcomes of the mean, where the weights correspond to $\kappa$ (virtual) measurements encoded in the prior and the $m$ (actual) measurements, i.e., the prior distribution $\mathcal{N}(\vec{\mu}_0, \bm{\Sigma} / \kappa)$ can be thought of as providing the information equivalent to $\kappa$ observations with sample mean $\vec{\mu}_0$.
Similarly, using Eq.~\eqref{eq:postdens_var}, the prior distribution $\mathcal{W}^{-1}(\nu \bm{\Sigma}_0, \nu)$ can be thought of as providing the information equivalent to $\nu$ observations with average squared deviation $\bm{\Sigma}_0$.

\newpage
\section{Derivation of multivariate generalization in detail}
\subsection{Moments}
\label{sec:moments}
We assume our data was drawn from a multivariate Gaussian with unknown mean and variance $(\vec{\mu}, \bm{\Sigma})$.
We probabilistically model these parameters $\bm{\theta}$ according to:
\begin{align*}
    \bm{\Sigma} &\sim \mathcal{W}^{-1}(\bm{\Psi}, \nu) \equiv \mathcal{W}^{-1}(\nu \bm{\Sigma}_0, \nu) \,, \\
    \vec{\mu} | \bm{\Sigma} &\sim \mathcal{N}(\vec{\mu}_0, \bm{\Sigma} / \kappa) \,,
\end{align*}
where $\mathcal{N}$ is a multivariate normal distribution
\begin{subequations}
\begin{align}
    \mathcal{N}(\vec{x}\, | \vec{\mu}, \bm{\Sigma})
    &= \frac{1}{\sqrt{(2 \pi)^n \, |\bm{\Sigma}|}} \exp\!\left\{ -\frac{1}{2} \, (\vec{x} - \vec{\mu})^\top \bm{\Sigma}^{-1} (\vec{x} - \vec{\mu}) \right\} \label{eq:gauss} \\
    &= \frac{1}{\sqrt{(2 \pi)^n \, |\bm{\Sigma}|}} \exp\!\left\{ -\frac{1}{2} \, \operatorname{tr}\!\left( (\vec{x} - \vec{\mu}) (\vec{x} - \vec{\mu})^\top \bm{\Sigma}^{-1} \right) \right\} \label{eq:gausstr}
\end{align}
\end{subequations}
and $\mathcal{W}^{-1}$ is an Inverse-Wishart distribution
\begin{equation*}
    \mathcal{W}^{-1}(\bm{\Sigma} | \bm{\Psi}, \nu)
    = \frac{1}{\Gamma_n(\nu/2)} \sqrt{\frac{1}{2^{\nu n}} \, \frac{|\bm{\Psi}|^\nu}{|\bm{\Sigma}|^{\nu + n + 1}}} \exp\!\left\{ -\frac{1}{2} \operatorname{tr}\!\left( \bm{\Psi} \bm{\Sigma}^{-1} \right) \right\} \,.
\end{equation*}
Using $\bm{\Sigma}_0$ rather than $\bm{\Psi}$ corresponds to parametrizing the distribution of $\bm{\Sigma}$ with an inverse $\chi^2$- rather than a $\Gamma^{-1}$-distribution in the univariate case.
This has the advantage of a clearer interpretation of $\bm{\Sigma}_0$, i.e., the prior distribution $\mathcal{W}^{-1}(\nu \bm{\Sigma}_0, \nu)$ can be thought of as providing the information equivalent to $\nu$ observations with average squared deviation $\bm{\Sigma}_0$.
Similarly, the prior distribution $\mathcal{N}(\vec{\mu}_0, \bm{\Sigma} / \kappa)$ can be though of as providing the information equivalent to $\kappa$ observations with sample mean $\vec{\mu}_0$, cf.\ Eqs.~\eqref{eq:postdens} and~\cite{gelman04}.

The prior joint distribution (a \texttt{NIW} distribution) factorizes and can be written as:
\begin{align*}
    p(\vec{\mu}, \bm{\Sigma}) &= p(\vec{\mu}) \times p(\bm{\Sigma}) \\
    &= \underbrace{\mathcal{N}(\vec{\mu} \, | \vec{\mu}_0, \bm{\Sigma} / \kappa) \times \mathcal{W}^{-1}(\bm{\Sigma} \, | \bm{\Psi}, \nu)}_{\texttt{NIW}} \,.
\end{align*}
The first order moments are then given by
\begin{subequations}
\begin{align}
    \langle \vec{\mu}\, \rangle_{\texttt{NIW}}
    &= \langle \vec{\mu}\, \rangle_{\mathcal{N}} = \vec{\mu}_0 \,, \\
    \langle \bm{\Sigma} \rangle_{\texttt{NIW}}
    &= \langle \bm{\Sigma} \rangle_{\mathcal{W}^{-1}} \nonumber \\
    &= \frac{1}{\nu - n - 1} \, \bm{\Psi} \nonumber \\
    &= \frac{\nu}{\nu - n - 1} \, \bm{\Sigma}_0
    \quad (\text{for } \nu > n + 1) \,.
\end{align}
Using these and
\begin{equation*}
        \langle \mu_i \, \mu_j \rangle_{\texttt{NIW}}
        = \langle \mu_i \, \mu_j \rangle_{\mathcal{N}}
        = \Sigma_{ij} / \kappa + \mu_{0,i} \, \mu_{0,j}
\end{equation*}
we find the variance of $\vec{\mu}$ being:
\begin{align}
    \operatorname{var}(\vec{\mu}\,)_{\texttt{NIW}}
    &= \begin{pmatrix}
        & \vdots & \\
        \dots & \langle \mu_i \, \mu_j \rangle_{\texttt{NIW}} & \dots \\
        & \vdots &
    \end{pmatrix} -
    \begin{pmatrix}
        & \vdots & \\
        \dots & \langle \mu_i \rangle_{\texttt{NIW}} \, \langle \mu_j \rangle_{\texttt{NIW}} & \dots \\
        & \vdots &
    \end{pmatrix} \nonumber \\
    &= \frac{1}{\kappa (\nu - n - 1)} \bm{\Psi} \nonumber \\
    &= \frac{\nu}{\kappa (\nu - n - 1)} \bm{\Sigma}_0 \quad (\text{for } \nu > n + 1) \,.
\end{align}
\end{subequations}
We note that in the univariate case, $n=1$, these relations become
\begin{subequations}
    \begin{align}
        \langle \mu \rangle &= \mu_0 \,, \\
        \langle \sigma^2 \rangle &= \beta / (\alpha - 1) \,, \\
        \operatorname{var}(\mu) &= \beta / (\kappa (\alpha - 1))
    \end{align}
\end{subequations}
for
\begin{align*}
    \sigma^2 &\sim \Gamma^{-1}(\alpha, \beta) \,, \\
    \mu | \sigma^2 &\sim \mathcal{N}(\mu_0, \sigma^2 / \kappa) \,,
\end{align*}
as expected.

\subsection{Model evidence}
\label{sec:appendix_modev}
Here we derive the posterior predictive or model evidence of a \texttt{NIW} distribution,
\begin{equation}
    \label{eq:niw}
    \operatorname{\texttt{NIW}} (\vec{\mu}, \bm{\Sigma} | \vec{\mu}_0, \bm{\Psi}, \kappa, \nu)
    = \mathcal{N}(\vec{\mu}\, | \vec{\mu}_0, \bm{\Sigma} / \kappa) \times \mathcal{W}^{-1}(\bm{\Sigma} | \bm{\Psi}, \nu) \,.
\end{equation}
where $\mathcal{N}(\vec{x})$ as well as $\mathcal{W}^{-1}(\bm{\Sigma})$ are proper normalized in $\vec{x}$ and $\bm{\Sigma}$, respectively, such that
\begin{subequations}
\begin{equation}
    \int \! \mathrm{d} \vec{x} \, \mathcal{N}(\vec{x} | \vec{\mu}, \bm{\Sigma}) = \int \! \mathrm{d} \vec{\mu} \, \mathcal{N}(\vec{x} | \vec{\mu}, \bm{\Sigma}) = \int \! \mathrm{d} \vec{\mu} \, \mathcal{N}(\vec{\mu} | \vec{x}, \bm{\Sigma}) = 1 \,, \label{eq:normnorm} \\
\end{equation}
\begin{equation}
    \int \! \mathrm{d} \bm{\Sigma} \, \mathcal{W}^{-1}(\bm{\Sigma} | \bm{\Psi}, \nu) = 1 \,. \label{eq:normiw}
\end{equation}
\end{subequations}

From Bayesian probability theory the model evidence is a marginal likelihood and, as such, defined as the likelihood of an observation $\vec{y}_i \in \mathbb{R}^n$ given the evidential distribution parameters $\mathfrak{m}=(\vec{\mu}_0, \bm{\Psi}, \kappa, \nu)$ and is computed by marginalizing over the likelihood parameter $\bm{\theta} = (\vec{\mu}, \bm{\Sigma})$, where $(\kappa, \nu) \in \mathbb{R}$, $(\vec{\mu}, \vec{\mu}_0) \in \mathbb{R}^n$ and $(\bm{\Sigma}, \bm{\Psi})$ are positive definite $\mathbb{R}^{n \times n}$ matrices:
\begin{equation}
    \label{eq:modev1}
    p(\vec{y}_i | \mathfrak{m}) = \frac{p(\vec{y}_i | \bm{\theta}, \mathfrak{m}) \, p(\bm{\theta} | \mathfrak{m})}{p(\bm{\theta} | \vec{y}_i, \mathfrak{m})} = \int \! \mathrm{d} \bm{\theta} \, p(\vec{y}_i | \bm{\theta}) \, p(\bm{\theta} | \mathfrak{m}) \,.
\end{equation}
In our case of placing a \texttt{NIW} evidential prior on a multivariate Gaussian likelihood function, i.e.,
\begin{equation*}
    p(\vec{y}_i | \mathfrak{m}) = \int \! \mathrm{d} \bm{\theta} \, \mathcal{N}(\vec{y}_i | \bm{\theta}) \,  \operatorname{\texttt{NIW}}(\bm{\theta} | \mathfrak{m}) \,,
\end{equation*}
an analytical solution exists and can be parametrized with a multivariate $t$-distribution with $\nu - n + 1$ DoF, cf.\ Sec.~\ref{sec:appendix_derivmodev}:
\begin{subequations}
\begin{align}
    \label{eq:modev21}
    p(\vec{y}_i | \mathfrak{m}) &= t_{\nu-n+1} \left( \vec{y}_i \middle| \vec{\mu}_0, \frac{1}{\nu - n + 1} \frac{1 + \kappa}{\kappa} \bm{\Psi} \right) \\
    &= \frac{\Gamma\!\left( \frac{\nu + 1}{2} \right)}{\Gamma\!\left( \frac{\nu - n + 1}{2} \right)} \sqrt{ \frac{\kappa^n}{(1 + \kappa)^n} \frac{1}{\pi^n | \bm{\Psi} |}}
    \times \left( 1 + \frac{\kappa}{1 + \kappa} (\vec{y}_i - \vec{\mu_0})^\top \bm{\Psi}^{-1} (\vec{y}_i - \vec{\mu}_0) \right)^{-\frac{\nu + 1}{2}} \nonumber \\
    &= \frac{\Gamma\!\left( \frac{\nu + 1}{2} \right)}{\Gamma\!\left( \frac{\nu - n + 1}{2} \right)} \sqrt{ \frac{\kappa^n}{(1 + \kappa)^n} \frac{1}{\pi^n |\bm{\Psi}|}}
    \times \left( \frac{\left| \bm{\Psi} + \frac{\kappa}{1 + \kappa} (\vec{y}_i - \vec{\mu_0}) (\vec{y}_i - \vec{\mu}_0)^\top \right|}{|\bm{\Psi}|} \right)^{-\frac{\nu + 1}{2}} \,, \label{eq:modev22}
\end{align}
\end{subequations}
where we used Sylvester's determinant theorem,
\begin{equation}
    \label{eq:sylvester}
    |\bm{\Psi} + c (\vec{\mu} - \vec{\mu}_0) (\vec{\mu} - \vec{\mu}_0)^\top| = |\bm{\Psi}| \left( 1 + c (\vec{\mu} - \vec{\mu}_0)^\top \bm{\Psi}^{-1} (\vec{\mu} - \vec{\mu}_0) \right)
\end{equation}
with $c \in \mathbb{R}$, to derive Eq.~\eqref{eq:modev22}.
Obviously, $p(\vec{y}_i | \mathfrak{m})$ on its own is not capable of defining $\mathfrak{m}$ unambiguously.
In particular, a fitting approach could be used to find $\nu$, $\vec{\mu}_0$ and the product $(1+\kappa)/\kappa \, \bm{\Psi}$ from data.
However, in order to disentangle the latter additional constraints have to be set, e.g., via an additional regularization of $\kappa$.

Using this result we can compute the negative log-likelihood loss $\mathcal{L}^{\text{NLL}}_i$ for sample $i$ as:
\begin{subequations}
\begin{align}
    \mathcal{L}^{\texttt{NLL}}_i &= - \log p(\vec{y}_i | \mathfrak{m}) \nonumber \\
    &= \log \Gamma\!\left( \frac{\nu - n + 1}{2} \right) - \log \Gamma\!\left( \frac{\nu + 1}{2} \right) \nonumber \\
    &\phantom{=} + \frac{n}{2} \log\!\left( \pi \, \frac{1 + \kappa}{\kappa} \right) - \frac{\nu}{2} \log |\bm{\Psi}| \nonumber \\
    &\phantom{=} + \frac{\nu + 1}{2} \log \left| \bm{\Psi} + \frac{\kappa}{1 + \kappa} (\vec{y}_i - \vec{\mu}_0) (\vec{y}_i - \vec{\mu}_0)^\top \right| \,, \label{eq:nll1}
\end{align}
or, alternatively, using a slightly different parametrization of $\mathfrak{m}$ with $\bm{\Psi} \equiv \nu \bm{\Sigma}_0$:
\begin{align}
    \mathcal{L}^{\texttt{NLL}}_i
    &= \log \Gamma\!\left( \frac{\nu - n + 1}{2} \right) - \log \Gamma\!\left( \frac{\nu + 1}{2} \right) \nonumber \\
    &\phantom{=} + \frac{n}{2} \log\!\left( \nu \pi \, \frac{1 + \kappa}{\kappa} \right) - \frac{\nu}{2} \log |\bm{\Sigma}_0| \nonumber \\
    &\phantom{=} + \frac{\nu + 1}{2} \log \left| \bm{\Sigma}_0 + \frac{\kappa}{1 + \kappa} \frac{1}{\nu} (\vec{y}_i - \vec{\mu_0}) (\vec{y}_i - \vec{\mu}_0)^\top \right| \,. \label{eq:nll2}
\end{align}
\end{subequations}

We note that in the univariate case, $n=1$, Eq.~\eqref{eq:modev21} becomes a non-standardized Student's $t$-distribution with $\nu$ DoF:
\begin{subequations}
\begin{align}
    p(y_i | \mathfrak{m}) &= \operatorname{St}_\nu \left( y_i \middle| \mu_0, \frac{1 + \kappa}{\kappa} \sigma_0^2 \right) \nonumber \\
    &= \frac{\Gamma\!\left( \frac{\nu + 1}{2} \right)}{\Gamma\!\left( \frac{\nu}{2} \right)} \sqrt{\frac{\kappa}{1 + \kappa} \frac{1}{\pi \nu \sigma_0^2}} \left( 1 + \frac{\kappa}{1 + \kappa} \frac{(y - \mu_0)^2}{\nu \sigma_0^2} \right)^{-\frac{\nu + 1}{2}}
\end{align}
and Eq.~\eqref{eq:nll1} reduces to
\begin{align}
    \mathcal{L}^{\texttt{NLL}}_i &= - \log p(\vec{y}_i | \mathfrak{m}) \nonumber \\
    &= \log \Gamma\!\left( \frac{\nu}{2} \right) - \log \Gamma\!\left( \frac{\nu + 1}{2} \right) \nonumber \\
    &\phantom{=} + \frac{1}{2} \log(\pi  / \kappa) - \frac{\nu}{2} \log(\nu \sigma_0^2 (1 + \kappa)) \nonumber \\
    &\phantom{=} + \frac{\nu + 1}{2} \log( \nu \sigma_0^2 (1 + \kappa) + \kappa (y_i - \mu_0)^2)
\end{align}
\end{subequations}
and thus reproduces the findings of~\cite{amini20} with $\nu = 2\alpha$, $\nu \sigma_0^2 = 2\beta$, $\kappa = v$ and $\mu_0 = \gamma$.
Similar to the multivariate case, $p(y, \mathfrak{m})$ on its own is not sufficient to define $\mathfrak{m}$ unambiguously.

\subsection{Analytical derivation of the model evidence}
\label{sec:appendix_derivmodev}
One way to derive Eq.~\eqref{eq:modev21} is, first, to use the fact that arguments can be shifted within the product of the two multivariate normal distributions $\mathcal{N}(\vec{y}_i | \vec{\mu}, \bm{\Sigma})$ and $\mathcal{N}(\vec{\mu}\, | \vec{\mu}_0, \bm{\Sigma} / \kappa)$,
\begin{align}
    \mathcal{N}(\vec{y}_i | \vec{\mu}, \bm{\Sigma}) \times \mathcal{N}(\vec{\mu}\, | \vec{\mu}_0, \bm{\Sigma} / \kappa)
    &= \mathcal{N}(\vec{\mu}\, | \vec{y}_i, \bm{\Sigma}) \times \mathcal{N}(\vec{\mu}\, | \vec{\mu}_0, \bm{\Sigma} / \kappa) \nonumber \\
    &= \mathcal{N}\! \left(\vec{y}_i \middle|\, \vec{\mu}_0, \frac{1 + \kappa}{\kappa} \bm{\Sigma} \right) \times \mathcal{N}\! \left(\vec{\mu} \,\middle|\, \frac{\vec{y}_i + \kappa \vec{\mu}_0}{1 + \kappa}, \frac{1}{1 + \kappa} \bm{\Sigma} \right) \,, \label{eq:trick1}
\end{align}
which we use to separate the integration parameter $\vec{\mu}_0$.
Secondly, the normal distribution \eqref{eq:gausstr} of a Normal-Inverse-Wishart distribution can be partially absorbed by the Inverse-Wishart distribution using a similar trick and Eq.~\eqref{eq:sylvester},
\begin{align}
    \operatorname{\texttt{NIW}} (\vec{\mu}, \bm{\Sigma} | \vec{\mu}_0, \bm{\Psi}, \kappa, \nu)
    &= \mathcal{N}(\vec{\mu}\, | \vec{\mu}_0, \bm{\Sigma} / \kappa) \times \mathcal{W}^{-1}(\bm{\Sigma} | \bm{\Psi}, \nu) \nonumber \\
    &= t_{\nu - n + 1}\!\left(\vec{\mu} \, \middle| \, \vec{\mu}_0, \frac{1}{\nu - n + 1} \bm{\Psi} / \kappa \right) \nonumber \\
    &\phantom{=} \times \mathcal{W}^{-1} \left( \bm{\Sigma} \middle| \bm{\Psi} + \kappa (\vec{\mu} - \vec{\mu}_0) (\vec{\mu} - \vec{\mu}_0)^\top, \nu + 1 \right) \,, \label{eq:trick2}
\end{align}
allowing for a straightforward integration of $\bm{\Sigma}$.
Using both we can evaluate integral \eqref{eq:modev1}:
\begin{align*}
    \int \! \mathrm{d} \bm{\theta} \, p(\vec{y}_i | \bm{\theta}) \, p(\bm{\theta} | \mathfrak{m})
    &= \iint \! \mathrm{d}\vec{\mu}_0 \, \mathrm{d} \bm{\Sigma} \, \mathcal{N}(\vec{y}_i | \vec{\mu}, \bm{\Sigma})
    \times \mathcal{N}(\vec{\mu} \, | \vec{\mu}_0, \bm{\Sigma} / \kappa) \times \mathcal{W}^{-1}(\bm{\Sigma} | \bm{\Psi}, \nu) \\
    &\overset{\eqref{eq:trick1}}{=} \int \mathrm{d} \bm{\Sigma} \left[ \int \! \mathrm{d} \vec{\mu} \, \mathcal{N}\! \left(\vec{\mu} \,\middle|\, \frac{\vec{y}_i + \kappa \vec{\mu}_0}{1 + \kappa}, \frac{1}{1 + \kappa} \bm{\Sigma} \right) \right] \\
    &\phantom{=} \times \mathcal{N}\! \left(\vec{y}_i \middle|\, \vec{\mu}_0, \frac{1 + \kappa}{\kappa} \bm{\Sigma} \right) \times \mathcal{W}^{-1}(\bm{\Sigma} | \bm{\Psi}, \nu) \\
    &\overset{\eqref{eq:normnorm}}{=} \int \! \mathrm{d} \bm{\Sigma} \, \mathcal{N}\! \left(\vec{y}_i \middle|\, \vec{\mu}_0, \frac{1 + \kappa}{\kappa} \bm{\Sigma} \right) \, \mathcal{W}^{-1}(\bm{\Sigma} | \bm{\Psi}, \nu) \\
    &\overset{\eqref{eq:niw}}{=} \int \! \mathrm{d} \bm{\Sigma} \, \operatorname{\texttt{NIW}} \!\left( \vec{y}_i, \bm{\Sigma} \middle| \vec{\mu}_0, \bm{\Psi}, \frac{\kappa}{1 + \kappa}, \nu \right) \\
    &\overset{\eqref{eq:trick2}}{=} t_{\nu - n + 1}\!\left(\vec{y}_i \middle| \, \vec{\mu}_0, \frac{1 + \kappa}{\kappa (\nu - n + 1)} \bm{\Psi} \right) \\
    &\phantom{=} \times \int \, \mathrm{d} \bm{\Sigma} \, \mathcal{W}^{-1} \left( \bm{\Sigma} \middle| \bm{\Psi} + \frac{\kappa}{1 + \kappa} (\vec{y}_i - \vec{\mu}_0)(\vec{y}_i - \vec{\mu}_0)^\top, \nu + 1 \right) \\
    &\overset{\eqref{eq:normiw}}{=} t_{\nu - n + 1}\!\left(\vec{y}_i \middle| \, \vec{\mu}_0, \frac{1 + \kappa}{\kappa (\nu - n + 1)} \bm{\Psi} \right) \,.
\end{align*}

\newpage
\section{Degeneration}
\label{sec:appendix_degeneration}
\begin{wrapfigure}{r}{.5\textwidth}
    \centering
    \includegraphics[scale=.8]{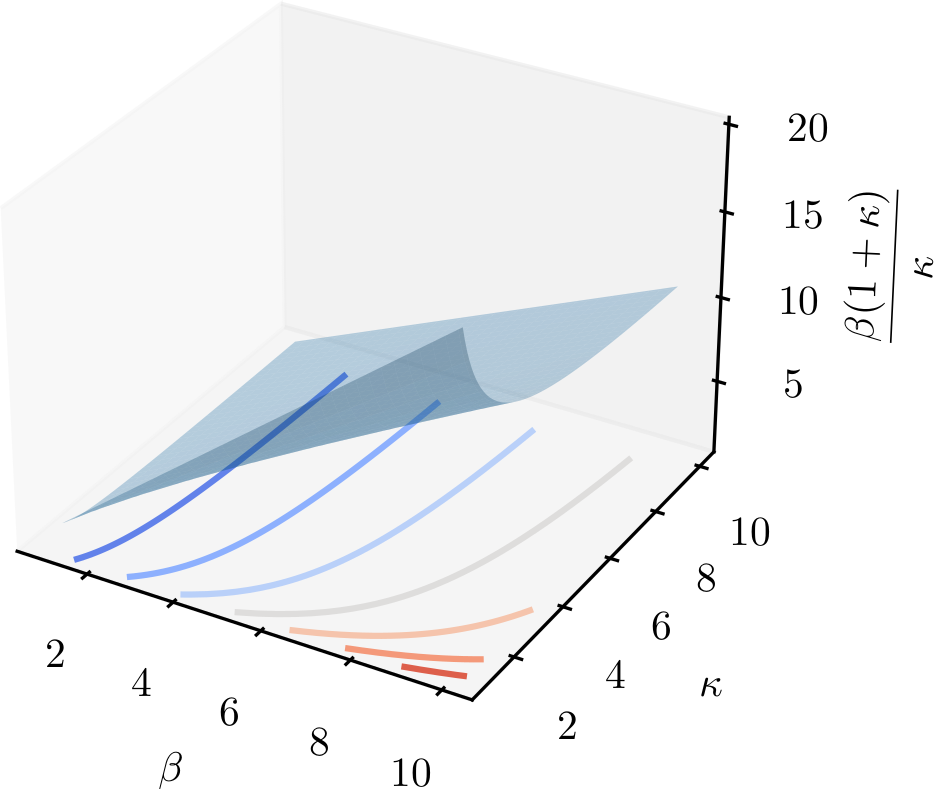}
    \caption{Parameters $\beta$ and $\kappa$, as well as their combination as used in the univariate loss function. Contour lines in the $\beta$-$\kappa$ plane indicate manifolds where the value of the loss function is constant.}
    \label{fig:degeneration}
\end{wrapfigure}
To understand the degeneration of the model evidence due to the ambiguity of $\kappa$ and $\bm{\Psi}$ it is helpful to look at the compact notation in Eq.~\eqref{eq:modev}:
The higher-order evidential distribution is projected by integrating out the nuisance parameters $\vec{\mu}$ and $\bm{\Sigma}$ and, in the univariate case, the four DoF of $\mathfrak{m}$ collapse into three DoF of a scaled Student's $t$-distribution.
Fitting this reduced set of DoF is not sufficient to recover all DoF of the evidential distribution.%
\footnote{This is similar to the example of fitting the distribution of the sum of two i.i.d.\ drawn samples from normal distributions with common mean $\mu$ but different variance $\sigma_1^2$ and $\sigma_2^2$. Here, the result will follow a normal distribution with mean $\mu$ and variance $\sigma_1^2 + \sigma_2^2$ where it is impossible to unfold $\sigma_1^2$ and $\sigma_2^2$ just by using the distribution of the sum -- the sum is a projection and one DoF is lost.}
The impact of this observation is that fitting the width of the $t$-distribution will not help to unfold $\kappa$ and $\bm{\Psi}$ and it is possible to find manifolds with different values of $\kappa$ and $\beta$ but with the same value for the loss function $\mathcal{L}_i^\text{NLL}$ as proposed in~\cite{amini20}.
In fact, $\kappa$ can be tuned such that for any given value of $\mathcal{L}_i^\text{NLL}$ a value for $\bm{\Psi}$ or $\beta$ can be find.
Since the inference of the parameters of NNs is achieved by minimizing a loss function, $\mathcal{L}_i^\text{NLL}$, which in our case solely depends on the product $\beta (1 + \kappa) / \kappa$ but not on $\beta$ and $\kappa$ individually, the contour lines in Fig.~\ref{fig:degeneration} correspond to configurations with constant value for $\mathcal{L}_i^\text{NLL}$ but different values for $\beta$ and $\kappa$ which we refer to as the \textit{degeneration}.
In tangible terms, each value of $\mathcal{L}_i^\text{NLL}$ maps to infinitely many aleatoric and epistemic uncertainties according to Eqs.~\eqref{eq:uncs}.
Hence, $\mathcal{L}_i^\text{NLL}$ on its own as proposed in~\cite{amini20}, is not sufficient to learn the parameters $\mathfrak{m}$.

\newpage
\section{Bias of fit parameters}
\label{sec:appendix_nubias}
\begin{figure}[htbp]
    \centering
    \includegraphics[scale=.8]{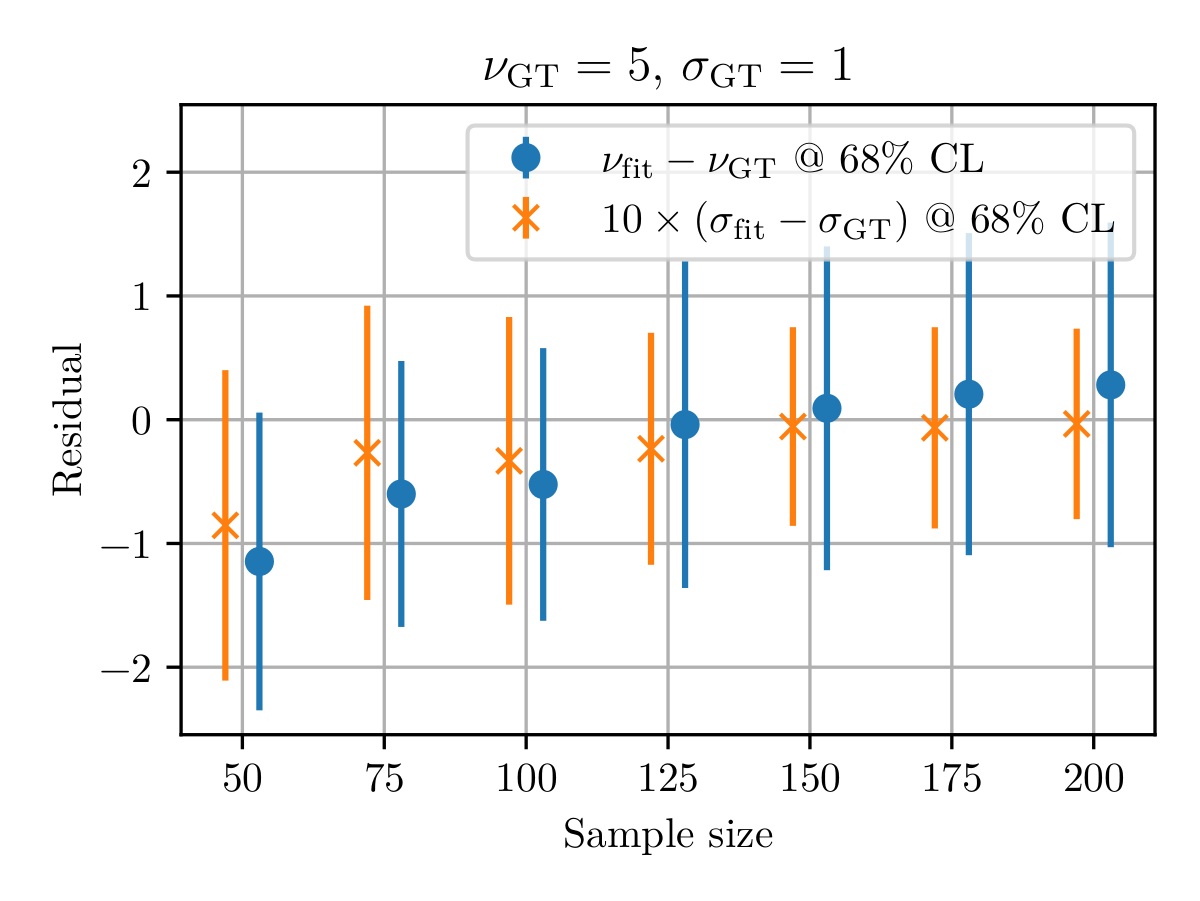}
    \caption{Residuals of the fitted parameter and the GT values used for generating the data. Here, a residual of zero corresponds to an unbiased estimate.}
    \label{fig:res}
\end{figure}

To study possible biases of fits with Student's $t$-distributions in the parameters $\nu$ and $\kappa$ we conduct a pseudo experiment where we generate data by drawing them i.i.d.\ from Student's $t$-distributions with fixed shape parameters (GT) and fit them on different sample sizes. 
For each sample size we evaluate 200 fits and indicate the values above (below) the 16\,\% (84\,\%) quantile in Fig.~\ref{fig:res} as error bars after subtracting the GT value.
This study shows a bias for $\nu$ and $\sigma$ which decreases if the sample size increases.

\newpage
\section{Multivariate data sample}
\label{sec:appendix_exp}
\begin{wrapfigure}{r}{.5\textwidth}
    \centering
    \includegraphics[scale=.8]{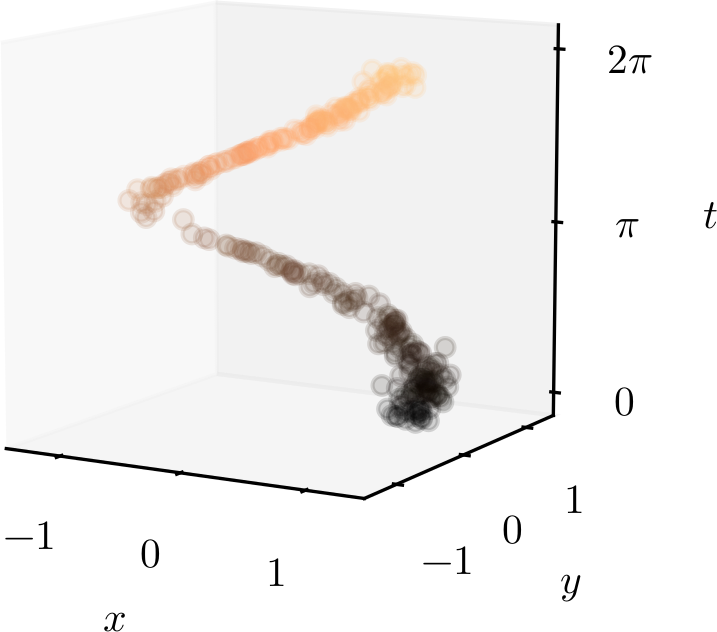}
    \caption{The distribution of our test data. The value of $t$ is color coded (see Fig.~\ref{fig:data} for a projection of the data into the $xy$-plane).}
    \label{fig:data_xyt}
\end{wrapfigure}
The data distribution used for the multivariate experiments outlined in Sec.~\ref{sec:multi} is shown in Fig.~\ref{fig:data_xyt}.
The NNs are trained to predict $(x, y) \in \mathbb{R}^2$ given $t \in \mathbb{R}$.
In total, 100 NNs of the same architecture are trained on the same data distribution.
The density of the data in $t$ varies and is maximal (minimal) at $t= \{ 0, 2\pi \}$ ($t=\pi$).
On top, Gaussian noise is added to $r$ but not $\varphi$ where we defined:
\begin{equation}
    \begin{pmatrix} r \\ \varphi \end{pmatrix} \overset{f}{\mapsto} \begin{pmatrix} x \\ y \end{pmatrix} = r \begin{pmatrix} \cos \varphi \\ \sin \varphi \end{pmatrix}.
\end{equation}
Using
\begin{equation}
    \operatorname{cov}(r, \varphi) = \begin{pmatrix} \sigma^2_r & 0 \\ 0 & 0 \end{pmatrix}
\end{equation}
the covariance of $x$ and $y$ can thus be estimated by using the Jacobian $\bm{J}_f$ of the mapping function $f$,
\begin{equation}
    \bm{J}_f = \begin{pmatrix} \cos \varphi & -r \sin \varphi \\ \sin \varphi & r \cos \varphi \end{pmatrix},
\end{equation}
yielding
\begin{equation}
    \operatorname{cov}(x, y) = \bm{J}_f \operatorname{cov}(r, \varphi)  \, \bm{J}_f^T = \sigma_r^2 \begin{pmatrix} \cos^2 \varphi & \sin \varphi \cos \varphi \\ \sin \varphi \cos \varphi & \sin^2 \varphi \end{pmatrix}.
\end{equation}
Note that this corresponds to a maximal correlation of $x$ and $y$,
\begin{equation}
    \operatorname{corr}(x, y) =
    \begin{cases}
        +1 & \text{for } 0 < t < \frac{\pi}{2} \,, \\
        -1 & \text{for } \frac{\pi}{2} < t < \pi \,, \\
        +1 & \text{for } \pi < t < \frac{3\pi}{2} \,, \\
        -1 & \text{for } \frac{3\pi}{2} < t <  2\pi \,.
    \end{cases}
\end{equation}

\begin{figure}[htbp]
    \centering
    \begin{subfigure}{.49\textwidth}
        \centering
        \includegraphics[width=\textwidth]{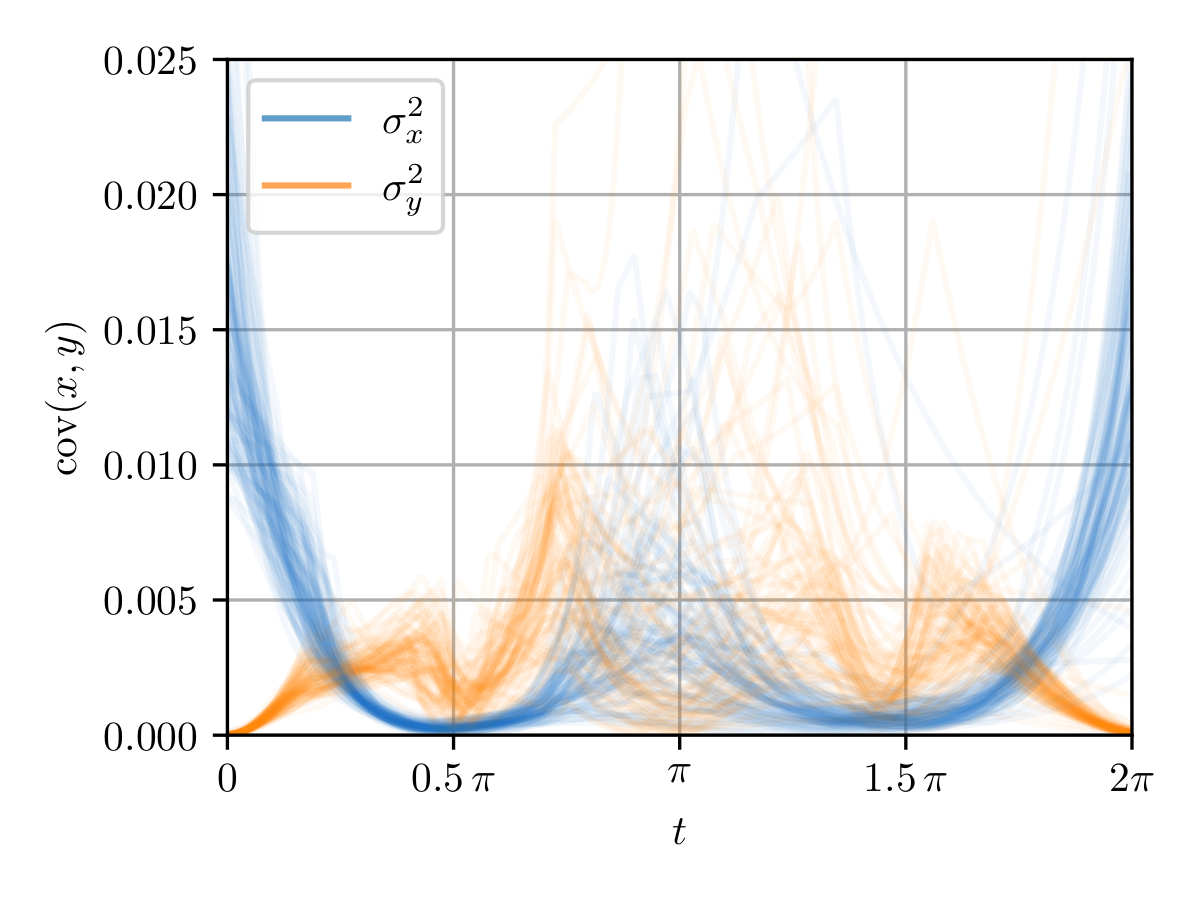}
    \end{subfigure}
    \begin{subfigure}{.49\textwidth}
        \centering
        \includegraphics[width=\textwidth]{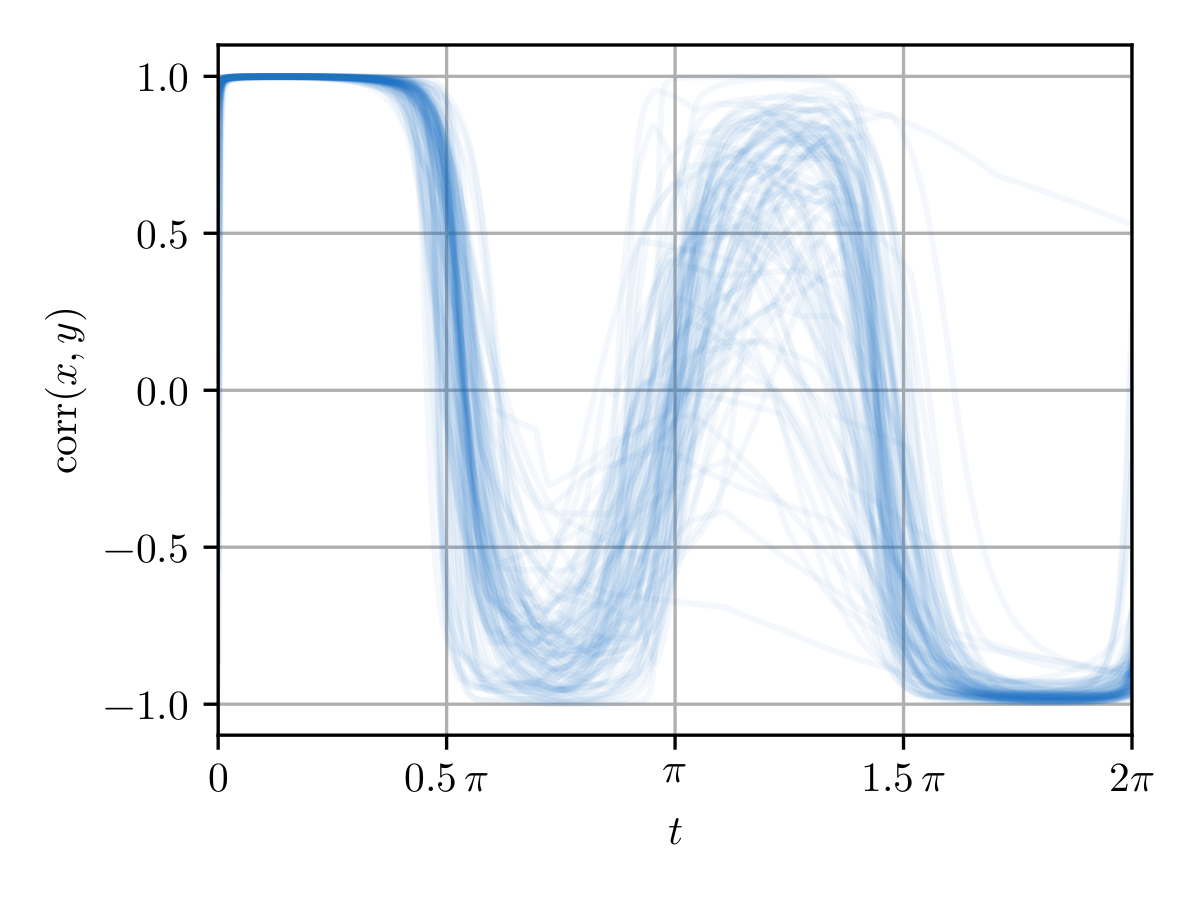}
    \end{subfigure}
    \caption{(Left) Overlayed covariance (up to a scaling constant) and (right) correlation of $x$ and $y$ as predicted by 100 NNs.}
    \label{fig:covcorr}
\end{figure}

The covariance for $x$ and $y$ as well as their correlation are also learned by NNs and respective predictions are shown in Fig.~\ref{fig:covcorr} (up to a scaling constant).
For regions with large statistic and constant correlation the expected behavior is well reproduced.
During transitions due to sign flips of the correlation at $t_\text{sf} = \{ 0.5\pi, \pi, 1.5\pi \}$ and in regions of large epistemic uncertainty the estimation of the covariance becomes unstable.
We find that if the sample size is increased (and thus the epistemic uncertainty is sufficiently suppressed) the  prediction of both values becomes stable for all values of $t$ except for small finite regions centered at $t_\text{sf}$.
\end{document}